\documentclass{article} 
\usepackage{iclr2026_conference,times}

\usepackage{amsmath,amsfonts,bm}

\def\eqref#1{equation~\ref{#1}}

\def\1{\bm{1}}

\DeclareMathAlphabet{\mathsfit}{\encodingdefault}{\sfdefault}{m}{sl}
\SetMathAlphabet{\mathsfit}{bold}{\encodingdefault}{\sfdefault}{bx}{n}

\DeclareMathOperator*{\argmin}{arg\,min}

\usepackage{hyperref}
\usepackage{url}
\usepackage{graphicx}
\usepackage{comment}
\usepackage{siunitx}
\usepackage{booktabs}
\usepackage{subcaption}

\title{TabINR: An Implicit Neural Representation Framework for Tabular Data Imputation}

\author{
Vincent Ochs\textsuperscript{1}, Florentin Bieder\textsuperscript{1}, Sidaty el Hadramy\textsuperscript{1}, Paul Friedrich\textsuperscript{1},\\
\textbf{Stephanie Taha-Mehlitz\textsuperscript{2}, Anas Taha\textsuperscript{1,3,4}, Philippe C. Cattin\textsuperscript{1}} \\
\textsuperscript{1} Department of Biomedical Engineering, Faculty of Medicine, University of Basel, 4123 Allschwil, Switzerland. \\
\textsuperscript{2} Clarunis, Department of Visceral Surgery, Basel, Switzerland. \\
\textsuperscript{3} Department of Visceral Surgery, Cantonal Hospital Basel-Land, Liestal, Switzerland. \\
\textsuperscript{4} Department of Surgery, East Carolina University, Brody School of Medicine, Greenville, NC, USA. \\
}

\iclrfinalcopy 
\begin{document}
\maketitle

\begin{abstract}
    Tabular data builds the basis for a wide range of applications, yet real-world datasets are frequently incomplete due to collection errors, privacy restrictions, or sensor failures. As missing values degrade the performance or hinder the applicability of downstream models, and while simple imputing strategies tend to introduce bias or distort the underlying data distribution, we require imputers that provide high-quality imputations, are robust across dataset sizes and yield fast inference. We therefore introduce \textsc{TabINR}, an auto-decoder based Implicit Neural Representation (INR) framework that models tables as neural functions. Building on recent advances in generalizable INRs, we introduce learnable row and feature embeddings that effectively deal with the discrete structure of tabular data and can be inferred from partial observations, enabling instance adaptive imputations without modifying the trained model. We evaluate our framework across a diverse range of twelve real-world datasets and multiple missingness mechanisms, demonstrating consistently strong imputation accuracy, mostly matching or outperforming classical (KNN, MICE, MissForest) and deep learning based models (GAIN, ReMasker), with the clearest gains on high-dimensional datasets.
\end{abstract}

\section{Introduction}

Tabular data remains one of the most common data formats across domains such as healthcare, finance, and the social sciences \citep{shwartz2022tabular}. 
In these fields, missing values are ubiquitous and can severely degrade the performance of downstream machine learning models. 
Poor handling of missingness not only reduces predictive accuracy but may also lead to biased decisions, with real-world consequences for applications such as medical diagnostics or financial risk assessment. 
These challenges make robust imputation a critical step for trustworthy tabular learning and data-driven decision making \citep{rubin1976inference}.

Unlike images, time series or text, tabular data exhibits unique characteristics that make imputation particularly difficult. 
First, it typically combines heterogeneous feature types such as continuous, categorical, and ordinal variables, which require different treatment. Moreover, many features exhibit complex nonlinear relationships and strong interdependencies, which must be taken into account \citep{shwartz2022tabular}. 
Second, tabular data usually lacks the spatial or sequential structure present in vision or language data. There is no inherent ordering of features across columns, and rows are often assumed to be independent. This removes many inductive biases that image or sequence models can exploit.
Third, many real-world tabular datasets are relatively small compared to their feature dimensionality, making it difficult for complex models to reliably generalize.

\begin{figure}
    \centering
    \includegraphics[width=\textwidth]{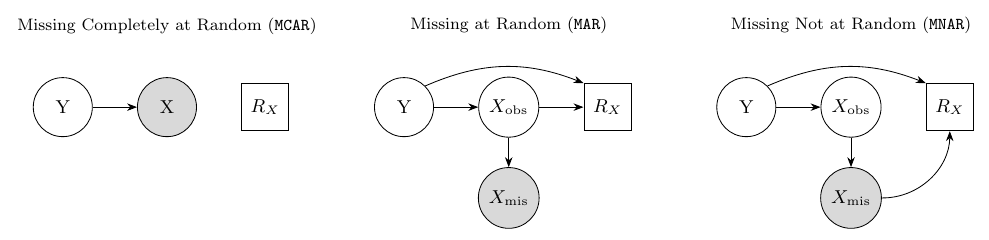}
    \caption{
Directed acyclic graphs illustrating common missing-data mechanisms for a partially observed variable $X$. White circles represent fully observed variables, gray circles represent unobserved or partially observed components, and squares denote the missingness indicators $R_X$. 
Arrows indicate causal or probabilistic dependence.}
    \label{fig:MissingDAG}
\end{figure}

An additional complication arises from the different mechanisms through which missing values occur \citep{rubin1976inference}. An overview of the three most common mechanisms is shown in \autoref{fig:MissingDAG}.
In the Missing Completely at Random (\texttt{MCAR}) setting, the probability of missingness is unrelated to either observed or unobserved data. While this assumption makes imputation relatively straightforward, it is also highly restrictive and rarely satisfied in practice.
In the more realistic Missing at Random (\texttt{MAR}) case, missingness depends on the observed variables but not on the missing ones themselves, requiring models to accurately capture conditional dependencies. 
The most challenging scenario is Missing Not at Random (\texttt{MNAR}), where the likelihood of missingness depends directly on the unobserved values. In this situation, unbiased estimation is generally impossible without imposing strong assumptions, explicitly modeling the missingness process, or incorporating domain-specific knowledge.
These factors make tabular imputation a central open problem for robust machine learning on structured data.

In this paper, we propose using Implicit Neural Representations (INRs) \citep{xie2022neural} for tabular data imputation. INRs represent data, in our case tables, as neural functions that map coordinates (e.g., row and column indices) to the corresponding values. We believe that such a neural data representation is a natural fit for imputation for several key reasons.
First, INRs can inherently fit data even when it is sparse or irregularly sampled. Once trained, the representation can be evaluated across the entire input domain, enabling the imputation of missing entries. Moreover, recent advances in generalizable INRs allow them to capture statistical regularities across datasets while still adapting to individual unseen rows through auto-decoder–style latent optimization \citep{park2019deepsdf}.
Unlike many conventional methods, INRs do not rely on strong distributional assumptions or arbitrary discretization choices. Instead, they learn a flexible representation capable of modeling complex dependencies directly from the data. Additionally, INRs are typically built using relatively simple Multilayer Perceptrons (MLPs), which results in a lightweight and fast solution.
Although INRs have been widely applied to images \citep{Sitzmann2020}, time series \citep{fons2022hypertime}, and 3D scenes \citep{mildenhall2021nerf}, their application to tabular data, together with the unique challenges and opportunities this entails, remains largely unexplored.
Our mian contributions are:
\begin{enumerate}
    \item We propose \textsc{TabINR}, a unified INR framework for tabular data imputation that leverages learnable row and feature embeddings to represent instances and variables in a shared latent space. 
    \item We introduce an auto-decoder-style test time latent optimization procedure that infers personalized embeddings for unseen instances from partial observations, enabling adaptability under sparse conditions. 
    \item We conduct extensive benchmarks against classical and deep learning baselines, showing that INR-based models achieve competitive or superior performance across diverse imputation tasks while offering a conceptually simple and memory-efficient alternative to GAN and transformer-based approaches.
\end{enumerate}
\subsection{Related Work}
\paragraph{Data Imputation Strategies} A variety of approaches have been proposed for imputing missing values in tabular datasets. 
Early strategies rely on simple heuristics such as mean or mode substitution \citep{little2019statistical}, expectation–maximization \citep{Jerez2010}, or matrix completion techniques \citep{Hastie2015}. 
More sophisticated statistical methods include MICE \citep{van1999flexible}, MissForest \citep{stekhoven2012missforest}, and MIRACLE \citep{Kyono2021}, which iteratively model conditional dependencies between features. 
These methods are efficient and widely used, but they either ignore complex nonlinear patterns or depend on correct model specification, while poorly scaling to large datasets \citep{white2011mice,shah2014comparison}.

Deep generative models take a different perspective by modeling the joint distribution of all features. 
GAN-based approaches such as GAIN \citep{Yoon2018GAIN} and GAMIN \citep{yoon2020gamin} treat imputation as a conditional generation problem, while VAE-based methods such as MIWAE \citep{Mattei2019} and HI-VAE \citep{nazabal2020handling} apply probabilistic inference to heterogeneous tabular data. 
Although expressive, these methods typically require large datasets and their adversarial or variational training objectives are often unstable and sensitive to hyperparameter choices. 
More recently, masked modeling and transformer-based architectures such as TabTransformer \citep{Huang2020}, FT-Transformer \citep{gorishniy2021revisiting}, SAINT \citep{somepalli2021saint}, and ReMasker \citep{du2024remasker} have demonstrated strong performance on imputation and downstream prediction. 
However, these models are often heavily overparameterized, incurring substantial computational costs and limiting their scalability, especially on smaller or noisier datasets.

\paragraph{Implicit Neural Representations}
Implicit Neural Representations (INRs) \citep{xie2022neural,essakine2024we} model data as continuous neural functions, that map from input coordinates, e.g., row and column indices, to the corresponding value. They have been applied to a large variety of different data modalities like sound \citep{Sitzmann2020}, images \citep{saragadam2023wire}, shapes \citep{park2019deepsdf}, videos \citep{chen2022videoinr}, or 3D scenes \citep{mildenhall2021nerf}, and have widely been adopted for a large variety of different tasks. 
Recent research has primarily focused on addressing the spectral bias \citep{rahaman2019spectral} inherent to MLPs typically used in INRs. Solutions include input embeddings such as Fourier Features \citep{tancik2020fourier}, specialized activation functions like SIREN \citep{Sitzmann2020}, Wire \citep{saragadam2023wire}, Gauss \citep{Ramasinghe2022}, HOSC \citep{serrano2024hosc}, and SINC \citep{saratchandran2024sampling}, as well as multi-resolution hash-grid encodings like InstantNGP \citep{muller2022instant}. Other approaches explore training strategies to identify the most informative samples \citep{kheradmand2024accelerating,tack2023learning} or techniques for optimal network initialization \citep{kania2024fresh}. 
While most INR research focuses on single-instance representations without cohort priors, generalization across multiple instances \citep{park2019deepsdf,dupont2022data,bieder2024modeling,friedrich2025medfuncta} has also been studied. For example, DeepSDF \citep{park2019deepsdf} introduced the auto-decoder framework, which jointly optimizes network weights and signal-specific latent vectors. This framework allows the network to learn patterns shared across different instances, while new signals can be efficiently fitted by optimizing only a new latent vector with the network weights kept frozen.

\section{Method}
\begin{figure}[ht]
    \centering
    \includegraphics[width=0.8\textwidth]{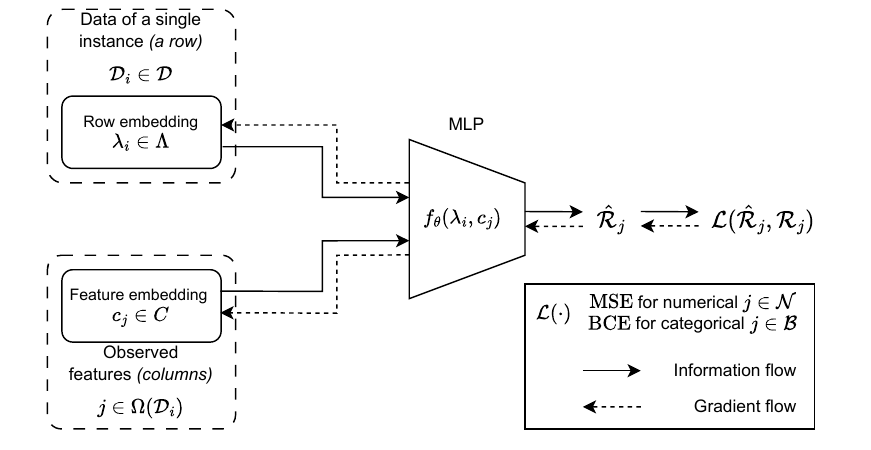}
    \caption{The proposed \textsc{TabINR} framework. During training, we jointly optimize the network $f_\theta$, the row embeddings $\Lambda$, as well as feature embeddings $C$. Once the network is trained, new instances can be added by only optimizing a new row embedding $\lambda_{\text{new}}$, keeping $f_\theta$ and $C$ fixed.}
    \label{fig:Fig1}
\end{figure}

We propose a unified INR framework, shown in \autoref{fig:Fig1}, for modeling tabular data. Rather than treating missingness as a nuisance, we parameterize the table as a neural function conditioned on row and feature embeddings. For each cell $(i,j)$, the model maps a learnable row embedding and a learnable feature embedding to a scalar value, enabling a single network to support imputation, downstream prediction, and instance-specific inference.

\subsection{TabINR}
Let $\mathcal{D} \in \overline{\mathbb{R}}^{n \times m}$ denote a tabular dataset with $n$ rows (instances) and $m$ columns (features), where $\overline{\mathbb{R}} = \mathbb{R} \cup \{\emptyset\}$ and $\emptyset$ denotes a missing entry. For each row $i$, the observed feature indices are denoted as 

\begin{equation}
\Omega(\mathcal{D}_{i:}) = \{ j \mid \mathcal{D}_{ij} \neq \emptyset \}.
\end{equation}

Instead of treating the table as a static array, we model each cell value $\mathcal{D}_{ij}$ as the output of a neural function conditioned on a pair of embeddings:
\begin{equation}
\hat{\mathcal{D}}_{ij} = f_\theta(\lambda_i, c_j),
\end{equation}
where:
\begin{itemize}
    \item $\lambda_i \in \Lambda$ is the row embedding representing instance $i$,
    \item $c_j \in C$ is the feature embedding representing column $j$,
    \item $f_\theta$ is a shared neural network (e.g.\ an MLP) that maps the embedding pair $(\lambda_i, c_j)$ to a scalar output $\hat{\mathcal{D}}_{ij}$.
\end{itemize}

During training, only observed pairs $(i,j)\in\Omega(\mathcal{D}_{i:})$ contribute to the loss. For numerical features, mean squared error (MSE) is used, while for categorical features, expanded via one-hot encoding (OHE), binary cross-entropy (BCE) with logits is applied. This formulation allows the same architecture to handle both feature types within a unified objective.

To identify a suitable network architecture, we conducted a grid search over key hyperparameters. The final configuration, indicated in \textbf{bold}, was selected based on the lowest 
average validation loss across all datasets. 
Specifically, we varied the row and feature embedding dimensions 
$\{16, \mathbf{32}, 64, 128, 256\}$, hidden layer widths 
$\{64, 128, \mathbf{256}, 512, 1024\}$, number of hidden layers 
$\{\mathbf{2}, 3, 4, 5, 10, 20\}$, dropout rates between $0.0$ and $\mathbf{0.1}$, 
activation functions $\{\text{ReLU}, \mathbf{\textbf{SIREN}}, \text{WIRE}, \text{SINC}\}$, 
and learning rates $\{10^{-2}, \mathbf{10^{-3}}, 10^{-4}\}$. 
We emphasize that this choice reflects a 
trade-off between stability and generalization across heterogeneous benchmarks, rather 
than the absolute optimum on any single dataset. Later ablation studies 
(\autoref{tab:Tab3}) demonstrate that larger models can achieve marginally better results 
on individual datasets.

\subsection{Training Strategy}

We define the set of observed entries as 
\begin{equation}
\mathcal{O} = \{(i,j) \mid \mathcal{D}_{ij} \neq \emptyset \}.
\end{equation}
Training is performed only on this subset, ensuring that missing entries never directly contribute to the objective. For each observed pair $(i,j)$, the model predicts
\begin{equation}
\hat{\mathcal{D}}_{ij} = f_\theta(\lambda_i, c_j),
\end{equation}
where $\lambda_i$ is the row embedding of instance $i$ and $c_j$ is the feature embedding of feature $j$.
To account for heterogeneous feature types, we use a mixed loss. Let $\mathcal{N}$ be the set of numerical features and let each original categorical feature $g$ be expanded into a one-hot group $\mathcal{B}_g=\{j_1,\dots,j_{K_g}\}$ of binary columns. Denote by $\mathcal{B}=\bigcup_g \mathcal{B}_g$ the set of all binary one-hot columns. The loss is
\begin{equation}
\mathcal{L}(\theta,\Lambda,C)
=\frac{1}{|\mathcal{O}|}\sum_{(i,j)\in\mathcal{O}}
\begin{cases}
\big(\hat{\mathcal{D}}_{ij}-\mathcal{D}_{ij}\big)^2, & j\in\mathcal{N},\\[6pt]
-\Big[\mathcal{D}_{ij}\,\log\sigma(\hat{\mathcal{D}}_{ij})+(1-\mathcal{D}_{ij})\log\big(1-\sigma(\hat{\mathcal{D}}_{ij})\big)\Big], & j\in\mathcal{B},
\end{cases}
\end{equation}
where $\sigma(\cdot)$ is the logistic sigmoid. Thus, for a categorical feature with $K_g$ categories, we compute one BCE term per one-hot component. Predicted logits are projected back to a valid category via $\arg\max$ within each one-hot group (winner-takes-all) during inference.
Optimization used Adam \citep{kingma2014adam} with cosine annealing learning rate scheduling and early stopping based on validation loss. During training, we applied random masking of \num{10}--\SI{70}{\percent} of entries to simulate missingness and evaluate reconstruction fidelity. 

\subsection{Test Time Adaptation via Latent Optimization}

At inference, we may encounter a new row $P \in \overline{\mathbb{R}}^m$ with only a subset of features observed, denoted by $\Omega(P)$. Since this row has no pretrained embedding, we introduce a fresh row embedding $\lambda_{\text{new}}$ (initialized randomly) and optimize it while keeping the network parameters $\theta$ and feature embeddings $\{c_j\}$ fixed:
\begin{equation}
\lambda_{\text{new}} 
= \argmin_{\lambda} 
\sum_{j \in \Omega(P)\cap\mathcal{N}} \big(f_\theta(\lambda, c_j) - P_j\big)^2
\;+\;
\sum_{j \in \Omega(P)\cap\mathcal{B}} \mathrm{BCEWithLogits}\ \!\big(f_\theta(\lambda, c_j), P_j\big).
\end{equation}

At inference, the row embedding is optimized to fit the observed features, and once this adaptation stabilizes, the missing entries can be imputed.

\section{Experiments and Results}

\paragraph{Datasets \& Preprocessing}

We benchmarked \textsc{TabINR} against commonly used imputers including mean/mode imputing, K-Nearest Neighbor (KNN), multiple imputation by chained equations (MICE)\citep{van1999flexible}, MissForest \citep{stekhoven2012missforest}, ReMasker \citep{du2024remasker}, and GAIN \citep{Yoon2018GAIN}. We evaluate our method on twelve publicly available real-world benchmarks spanning different domains, with datasets drawn from the UCI Machine Learning Repository \citep{Dua:2017}. These datasets vary widely in sample size, dimensionality, and feature composition, covering low- and high-dimensional features as well as mixed numerical and categorical variables. A detailed summary of dataset statistics, including size and number of features, is provided in the Appendix (\autoref{tab:Tab4}). This enables a systematic evaluation of \textsc{TabINR} across heterogeneous tabular data.

Preprocessing follows a consistent pipeline. Column types are inferred (or taken from dataset metadata) and numerical features are retained as real-valued, while categorical features are expanded via OHE. During training, numerical features are min–max scaled inside the loop (for both TabINR and GAIN), while OHE columns remain binary. At inference, categorical predictions are projected back to valid one-hot vectors using a winner-takes-all heuristic.  

\textsc{TabINR} is implemented in PyTorch. Row and feature embeddings are initialized from a standard normal distribution and optimized jointly with the shared MLP, which uses SIREN activation functions. Optimization employs Adam with cosine annealing learning-rate scheduling and early stopping on a validation split. All experiments were conducted on an NVIDIA A100 GPU (\SI{40}{\giga\byte}). Competing baselines were configured in line with prior work \citep{Yoon2018GAIN,Mattei2019,Hastie2015,jarrett2022hyperimpute} to ensure comparability.

\paragraph{Missingness Synthetization}

As the UCI benchmarks are fully observed, we synthetically introduce missing values at varying rates $p_{\text{miss}}$ under three standard mechanisms, shown in \autoref{fig:MissingDAG}. In the \texttt{MCAR} setting, entries are independently masked according to a Bernoulli distribution, yielding uniform missingness across the table. In the \texttt{MAR} setting, a subset of features is designated as always observed, while the others are masked using a logistic model conditioned on the observed subset. \texttt{MNAR}, extends \texttt{MAR} by additionally applying Bernoulli masking to values left unmasked, creating missingness that depends directly on the underlying data distribution.  

To ensure consistency with established practice, we adapt the missingness mechanisms implementation from HyperImpute \citep{jarrett2022hyperimpute}, which has been widely used in prior imputation studies \citep{Yoon2018GAIN,Mattei2019,Hastie2015}. The fully observed, OHE-expanded matrices serve as ground truth for evaluation, while the incomplete tables and masks define the observed entries available to the model during training and the held-out entries for evaluation. 

\begin{figure}[t]
    \centering
    \includegraphics[width=1.0\textwidth]{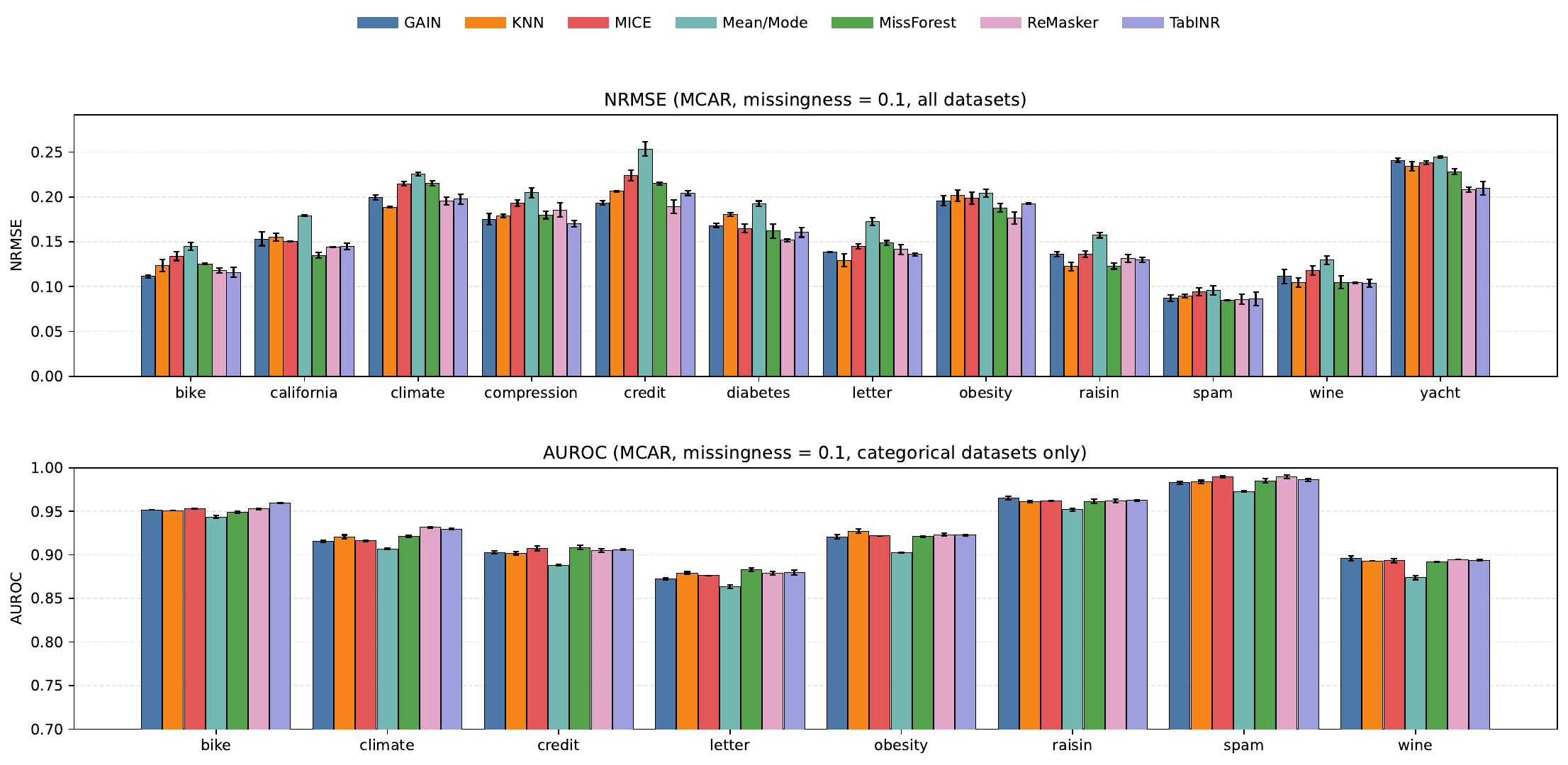}
\caption{Overall performance of TabINR and six baselines on \num{12} benchmark datasets under \texttt{MCAR} with
\num{0.1} missingness ratio. The results are shown as the mean and standard deviation of RMSE, and
AUROC scores (AUROC is only applicable to datasets with classification tasks).}
\label{fig:Fig2}
\end{figure}

\subsection{Benchmark: Imputation}

We train \textsc{TabINR} on each dataset and evaluate its ability to reconstruct both numerical and categorical variables under controlled missingness. For numerical features, performance is quantified using normalized root mean squared error (NRMSE), where per-feature RMSE is normalized by the feature’s standard deviation and averaged across variables. For categorical features, evaluation is based on the area under the receiver operating characteristic curve (AUROC).

Across all three missingness mechanisms (\texttt{MCAR}, \texttt{MAR}, \texttt{MNAR}) and missingness rates ranging from \SI{10}{\percent} to \SI{70}{\percent}, \textsc{TabINR} consistently ranks among the top-performing imputers and frequently achieves the best accuracy on high-dimensional datasets such as \textit{bike} and \textit{spam}. Under \texttt{MCAR} conditions, differences between methods remain modest, yet \textsc{TabINR} typically matches or exceeds the strongest classical baselines. Under \texttt{MAR} and especially \texttt{MNAR}, the performance spread widens: KNN and MICE degrade substantially due to their sensitivity to missingness patterns, while \textsc{TabINR} maintains stable reconstruction accuracy. Among competing methods, ReMasker is the most competitive deep baseline, while MissForest remains strong on smaller or predominantly categorical datasets. Nevertheless, \textsc{TabINR} generally delivers lower NRMSE for numerical features and comparable or higher AUROC for categorical features across the majority of benchmarks.

As expected, performance worsens for all methods as the missingness ratio increases, but \textsc{TabINR} degrades gracefully. Even at \SI{70}{\percent} missingness, it remains within the top three methods across nearly all datasets, with only occasional wins by ReMasker or MissForest on structured, lower-dimensional tables. These findings align with our sensitivity analysis, where \textsc{TabINR} benefited most from larger sample sizes and higher feature counts, whereas iterative (MICE) and local (KNN) approaches rapidly lost ground as sparsity and dimensionality increased. Overall, the results demonstrate that \textsc{TabINR} achieves robust numerical imputations (low NRMSE) and strong categorical reconstruction (high AUROC), with the clearest advantages observed in non-\texttt{MCAR} and high-dimensional datasets. Results are summarized in \autoref{fig:Fig2} and reported in detail in the Appendix (\autoref{fig:Fig5}--\autoref{fig:Fig15}).

\begin{figure}[t]
    \centering
    \includegraphics[width=0.9\textwidth]{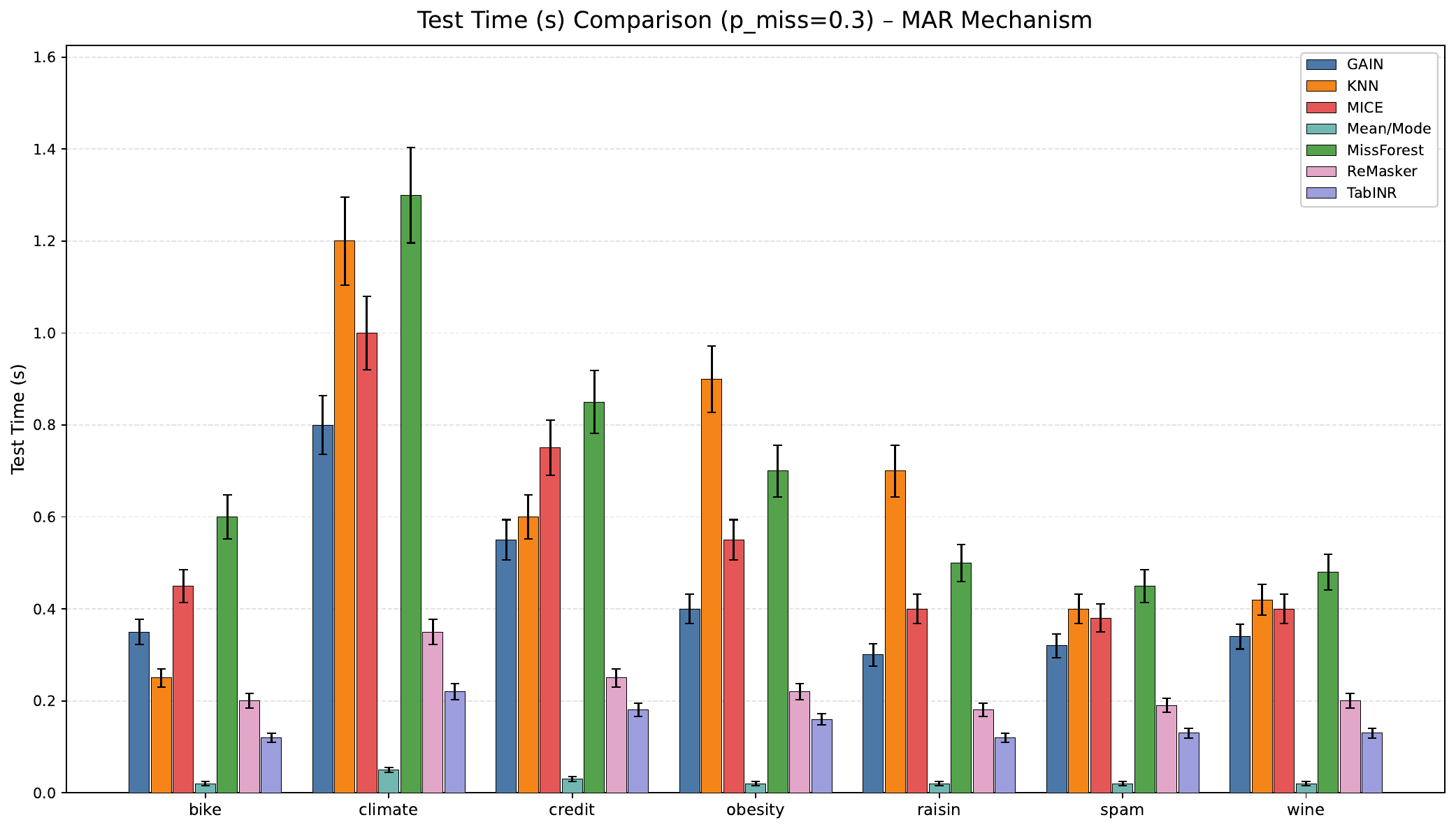}
    \caption{Inference-time comparison across imputers (lower is better). Bars show mean seconds per dataset over \num{5} runs; error bars denote $\pm$\num{1}$\,\mathrm{SD}$. While Mean/Mode is trivially fastest, \textsc{TabINR} and ReMasker achieve sub-\SI{0.25}{\second} inference once trained, whereas iterative baselines (KNN, MICE, MissForest) are markedly slower on higher-dimensional datasets. Results shown for \texttt{MAR} with $p_{\text{miss}}=0.3$; trends are consistent under \texttt{MCAR}/\texttt{MNAR}.}

    \label{fig:Fig3}
\end{figure}

\subsubsection{Inference Time Efficiency}

In addition to reconstruction quality, we benchmark test time efficiency across methods. Classical iterative approaches such as MICE, MissForest, and KNN exhibit the highest runtimes, frequently exceeding one second per dataset and scaling poorly with larger feature spaces (e.g., \textit{climate}, \textit{credit}). In contrast, simple heuristics such as mean/mode imputation are trivially fastest, as they involve no training or iterative optimization. Generative approaches such as ReMasker, as well as our proposed \textsc{TabINR}, achieve efficient inference once trained, requiring only \num{0.1}–\num{0.2} seconds on most datasets. This efficiency stems from the embedding-based formulation: imputations are performed through lightweight forward passes, rather than repeated regressions or tree growth. Inference-time comparisons are shown in \autoref{fig:Fig3}. While results reported in the main text focus on \texttt{MAR} with \SI{30}{\percent} missingness for clarity, the same trends hold under \texttt{MCAR} and \texttt{MNAR}.

\subsubsection{Permutation Robustness}

A potential concern when applying implicit neural representations to tabular data is that row and column indices lack inherent ordering, raising the question of whether performance depends on dataset permutation. \textsc{TabINR} solves this issue by not relying on absolute positional encodings: each row and each feature is associated with a learnable embedding, and the model reconstructs entries by combining these embeddings. Consequently, permuting rows or columns simply permutes the associated embeddings without altering the learned mapping. To verify this empirically, we performed experiments in which both rows and columns were permuted before training, confirming that NRMSE and AUROC remained stable. \autoref{tab:Tab1} reports the results, demonstrating that \textsc{TabINR} is invariant to dataset permutations and does not exploit any hidden spatial structure. Hyperparameters for these experiments are provided in the Appendix \autoref{tab:Tab5}.

\begin{table}[h]
\centering
\caption{Permutation robustness on \textit{letter} (\texttt{MCAR}, $p_{\text{miss}}{=}0.3$). 
Each row permutes feature indices before training. Performance is stable across shuffles.}
\label{tab:Tab1}
\small
\begin{tabular}{c|c|cc}
\toprule
\textbf{Sample} & \textbf{Permutation (indices \texttt{0}--\texttt{15})} & \textbf{NRMSE} ($\downarrow$) & \textbf{AUROC} ($\uparrow$) \\
\midrule
baseline & \texttt{0,1,2,3,4,5,6,7,8,9,10,11,12,13,14,15} & \num{0.128} & \num{0.851} \\
\num{1} & \texttt{7,3,12,1,9,6,14,0,5,10,2,8,15,4,11,13} & \num{0.128} & \num{0.852} \\
\num{2} & \texttt{5,14,2,8,11,0,10,3,12,7,1,9,6,13,15,4} & \num{0.129} & \num{0.850} \\
\num{3} & \texttt{9,0,7,12,3,10,15,6,1,14,8,5,2,11,4,13} & \num{0.127} & \num{0.853} \\
\num{4} & \texttt{11,6,1,15,8,3,4,12,14,2,9,0,13,10,5,7} & \num{0.128} & \num{0.849} \\
\num{5} & \texttt{2,9,5,13,0,8,10,1,7,4,12,15,3,14,6,11} & \num{0.128} & \num{0.852} \\
\num{6} & \texttt{4,8,14,6,12,11,9,2,3,7,13,1,10,5,15,0} & \num{0.129} & \num{0.850} \\
\num{7} & \texttt{10,5,0,7,1,12,3,9,6,13,15,2,8,11,4,14} & \num{0.128} & \num{0.851} \\
\num{8} & \texttt{13,2,11,4,7,1,8,15,10,0,6,14,5,3,12,9} & \num{0.127} & \num{0.852} \\
\bottomrule
\end{tabular}
\end{table}

\subsection{Downstream Classification}

We conducted a downstream classification task to evaluate the practical utility of imputations in predictive pipelines. We simulated missingness for each dataset with categorical targets, performed full-column imputation of the target variable using different imputation methods, and then trained XGBoost classifiers \citep{chen2016xgboost} on the completed datasets. This setup reflects a realistic use case in which imputation quality directly impacts the performance of a subsequent model. The preparation process mirrors the imputation benchmark. The fully imputed datasets then served as input to XGBoost, with AUROC as the evaluation metric. Results were averaged across four missingness ratios (\SI{10}{\percent}, \SI{30}{\percent}, \SI{50}{\percent}, and \SI{70}{\percent}) to ensure robustness. \autoref{tab:Tab2} reports AUROC scores on five UCI datasets with categorical targets. \textsc{TabINR} combined with XGBoost achieved the best performance on \textit{obesity} and \textit{spam}, while ReMasker performed slightly stronger on \textit{letter}. On \textit{raisin} and \textit{credit}, KNN-based imputation provided competitive results. We additionally evaluated the performance of \textsc{TabINR} as a classifier, by treating the target label as the variable to be imputed.

\begin{table}[h]
\centering
\caption{Downstream classification (full-column imputation of the target) measured by AUROC (mean $\pm$ std). Best result per dataset is \textbf{bold}.}
\label{tab:Tab2}
\resizebox{\textwidth}{!}{
\begin{tabular}{l l|c|c|c|c|c}
\toprule
\textbf{Pipeline} &  & \textbf{credit} & \textbf{letter} & \textbf{obesity} & \textbf{raisin} & \textbf{spam} \\
\midrule
Original Data (fully observed) & +XGBoost & $.835 \pm .008$ & $.930 \pm .006$ & $.940 \pm .006$ & $.840 \pm .008$ & $.915 \pm .006$ \\
\textsc{TabINR} Imputation & +XGBoost & $.848 \pm .007$ & $.927 \pm .006$ & $\mathbf{.947 \pm .005}$ & $.852 \pm .007$ & $\mathbf{.922 \pm .005}$ \\
KNN Imputation & +XGBoost & $\mathbf{.852 \pm .007}$ & $.922 \pm .007$ & $.942 \pm .006$ & $\mathbf{.855 \pm .007}$ & $.918 \pm .006$ \\
MissForest Imputation & +XGBoost & $.850 \pm .007$ & $.925 \pm .006$ & $.945 \pm .005$ & $.853 \pm .007$ & $.920 \pm .005$ \\
MICE Imputation & +XGBoost & $.847 \pm .007$ & $.921 \pm .007$ & $.943 \pm .006$ & $.852 \pm .007$ & $.917 \pm .006$ \\
GAIN Imputation & +XGBoost & $.845 \pm .007$ & $.919 \pm .007$ & $.941 \pm .006$ & $.851 \pm .007$ & $.916 \pm .006$ \\
ReMasker Imputation & +XGBoost & $.851 \pm .007$ & $\mathbf{.932 \pm .006}$ & $.946 \pm .005$ & $.854 \pm .007$ & $.921 \pm .005$ \\
Direct \textsc{TabINR} classifier &  & $.820 \pm .009$ & $.910 \pm .008$ & $.930 \pm .007$ & $.830 \pm .008$ & $.905 \pm .007$ \\
\bottomrule
\end{tabular}}
\end{table}

Overall, these findings confirm that imputations produced by \textsc{TabINR} translate into strong downstream predictive performance. While no single method dominates across all datasets, \textsc{TabINR} consistently provides competitive or superior results, demonstrating that INR-based imputations are not only accurate in reconstruction but also effective for downstream classification tasks.

\subsection{Ablation Study}

To better understand the robustness of all comparing methods, we conduct multiple ablation studies by varying the dataset size, feature dimensionality, and missingness ratio. Several consistent trends emerged: (a) performance of all methods improved with larger datasets, but TabINR and ReMasker gained a more pronounced advantage with larger data sizes; (b) as the number of features increased, TabINR maintained stable performance, while baseline methods such as KNN and MICE degraded; and (c) under higher missingness ratios (greater than \num{0.5}), performance differences between models narrowed, with TabINR and ReMasker remaining the most resilient. These results highlight the scalability of INR-based approaches and their particular strength on challenging, high-dimensional settings. This can be observed in \autoref{fig:Fig4}.
Additional ablation experiments across the other missingness mechanisms can be found in the Appendix (\autoref{fig:Fig16}--\autoref{fig:Fig26}).
\begin{figure}[ht]
    \centering
    \includegraphics[width=1.0\textwidth]{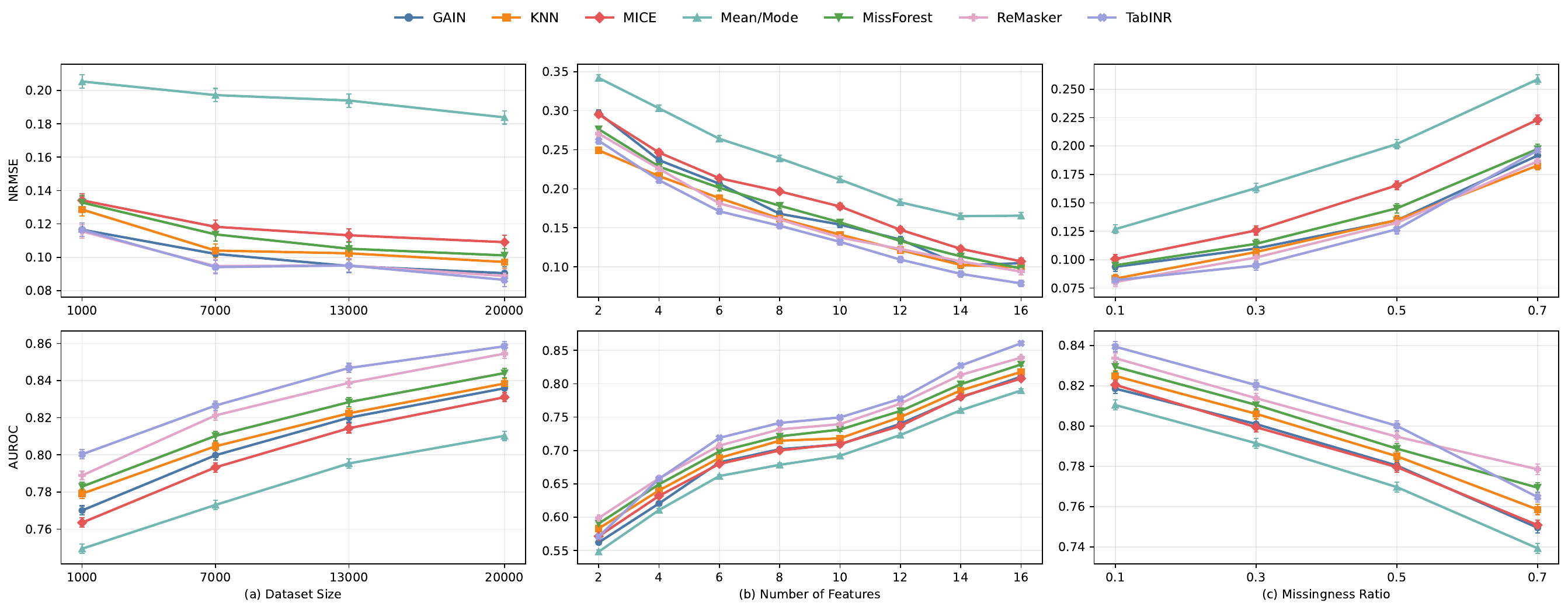}
\caption{Sensitivity analysis of TabINR on the \textit{letter} dataset under \texttt{MCAR} scenarios.
The results are shown in terms of RMSE and AUROC, with the scores measured with respect to (a) the dataset size,
(b) the number of features, and (c) the missingness ratio. The default setting is as follows: dataset size = \num{20000},
number of features = \num{16}, and missingness ratio = \num{0.1}.}
\label{fig:Fig4}
\end{figure}

\autoref{tab:Tab3} reports the ablation over depth, latent dimension, width, and activation on \textit{letter} (\texttt{MCAR}, $p_{\text{miss}}{=}0.1$). Performance improves with capacity and then plateaus: best NRMSE occurs around \num{10} layers with \num{512}–\num{1024} units and a \num{256}-d latent ($\approx$ \num{0.125}); SIREN achieves the highest AUROC (\num{0.864}), and HOSC the lowest NRMSE (\num{0.125}). This trend is dataset-specific, and in our main results we adopt the global defaults from Section~2.1 (Table~\ref{tab:Tab5}) to ensure robustness and comparability across benchmarks, rather than per-dataset peak tuning.

\begin{table}[ht]
\centering
\caption{Ablation study of \textsc{TabINR} on the \textit{letter} dataset (\texttt{MCAR}, $p_{\text{miss}}=0.1$). 
For comparability, we define a separate default configuration here 
(depth = \num{4}, latent dimension = \num{64}, hidden units = \num{256}, activation = SIREN), 
which differs from the global defaults in \autoref{tab:Tab5}. 
This setup serves only as a controlled baseline for sensitivity analysis. 
Results show that performance improves with additional capacity (depth, width, latent size), 
although gains plateau beyond a certain point and do not consistently generalize across all datasets.}
\label{tab:Tab3}

\begin{subtable}[t]{0.48\textwidth}
    \centering
    \caption{Layers}
    \begin{tabular}{S[table-format=3.0]|c|c}
        \toprule
        \textbf{Layers} & \textbf{NRMSE} ($\downarrow$) & \textbf{AUROC} ($\uparrow$) \\
        \midrule
        2   & \num{0.134} & \num{0.843} \\
        3   & \num{0.130} & \num{0.846} \\
        4   & \num{0.128} & \num{0.850} \\
        5   & \num{0.126} & \num{0.852} \\
        10  & \num{0.125} & \num{0.854} \\
        20  & \num{0.127} & \num{0.853} \\
        \bottomrule
    \end{tabular}
\end{subtable}
\hfill
\begin{subtable}[t]{0.48\textwidth}
    \centering
    \caption{Latent dimension}
    \begin{tabular}{S[table-format=3.0]|c|c}
        \toprule
        \textbf{Latent Dim} & \textbf{NRMSE} ($\downarrow$) & \textbf{AUROC} ($\uparrow$) \\
        \midrule
        16   & \num{0.139} & \num{0.836} \\
        32   & \num{0.133} & \num{0.843} \\
        64   & \num{0.128} & \num{0.851} \\
        128  & \num{0.126} & \num{0.854} \\
        256  & \num{0.125} & \num{0.855} \\
        \bottomrule
    \end{tabular}
\end{subtable}

\vspace{1em}

\begin{subtable}[t]{0.48\textwidth}
    \centering
    \caption{Units per hidden layer}
    \begin{tabular}{S[table-format=4.0] |c|c}
        \toprule
        \textbf{Units} & \textbf{NRMSE} ($\downarrow$) & \textbf{AUROC} ($\uparrow$) \\
        \midrule
        64    & \num{0.137} & \num{0.840} \\
        128   & \num{0.132} & \num{0.846} \\
        256   & \num{0.129} & \num{0.850} \\
        512   & \num{0.126} & \num{0.853} \\
        1024  & \num{0.126} & \num{0.854} \\
        \bottomrule
    \end{tabular}
\end{subtable}
\hfill
\begin{subtable}[t]{0.48\textwidth}
    \centering
    \caption{Activation functions}
    \begin{tabular}{l|c|c}
        \toprule
        \textbf{Activation} & \textbf{NRMSE} ($\downarrow$) & \textbf{AUROC} ($\uparrow$) \\
        \midrule
        ReLU   & \num{0.129} & \num{0.848} \\
        SIREN  & \num{0.127} & \num{0.864} \\
        Wire   & \num{0.126} & \num{0.855} \\
        HOSC   & \num{0.125} & \num{0.856} \\
        \bottomrule
    \end{tabular}
\end{subtable}

\end{table}

\section{Conclusion}

Our study demonstrates that Implicit Neural Representations (INRs) offer a simple yet flexible framework for imputing missing values in tabular data. By parameterizing entries with learnable row and feature embeddings and enabling instance-level adaptation through test time optimization, \textsc{TabINR} bridges the gap between classical imputers and recent deep learning based approaches.

While our results are promising, several limitations remain. First, our experiments focused on moderate-scale benchmarks with synthetically induced missingness. Applying \textsc{TabINR} under real-world settings with more complex missingness patterns and the need for integrating domain knowledge, remains unexplored. Second, we employed a single global default configuration across datasets to ensure comparability. This favors stability but likely underestimates the model’s best-case performance. Finally, although we compared against strong classical and generative baselines, we did not exhaustively benchmark against the latest transformer-based classifiers trained end-to-end.

For future work, we aim to extend \textsc{TabINR} to handle non-random missingness, scale to larger datasets, and incorporate automated hyperparameter adaptation. We also plan to integrate \textsc{TabINR} into multimodal pipelines for jointly modeling tabular data with images or text.

More broadly, our work suggests that INRs can serve as a unifying lens for tabular learning. By framing missing value imputation as a continuous representation learning problem, \textsc{TabINR} opens new directions for bringing the benefits of implicit representations to core challenges in tabular data analysis.

\section{ACKNOWLEDGMENTS}
This work was funded by the grant A1813455-SY from Medtronic and by the Vontobel foundation with the number: 0132/2024. Any opinions, findings, and conclusions
or recommendations expressed in this material are those of the author(s) and do not necessarily reflect
the views of the funder.

\bibliography{iclr2026_conference}
\bibliographystyle{iclr2026_conference}

\appendix
\section{Appendix}

\begin{table}[h]
\centering
\caption{Characteristics of the datasets used in our experiments.}
\label{tab:Tab4}
\begin{tabular}{lrrl}
\toprule
\textbf{Dataset} & \textbf{Dataset Size} & \textbf{\# Features} & \textbf{Task} \\
\midrule
(California) Housing               & \num{20640} & \num{9}  & Regression \\
(Climate) Model Simulation Crashes & \num{540}   & \num{18} & Classification \\
Concrete (Compressive) Strength    & \num{1030}  & \num{9}  & Regression \\
(Diabetes)                         & \num{442}   & \num{10} & Regression \\
Estimation of (Obesity) Levels     & \num{2111}  & \num{17} & Classification \\
(Credit) Approval                  & \num{690}   & \num{15} & Classification \\
(Wine) Quality                     & \num{1599}  & \num{12} & Classification \\
(Raisin)                           & \num{900}   & \num{8}  & Classification \\
(Spam) Base                        & \num{4601}  & \num{57} & Classification \\
(Bike) Sharing Demand              & \num{8760}  & \num{14} & Classification \\
(Letter) Recognition               & \num{20000} & \num{16} & Classification \\
(Yacht) Hydrodynamics              & \num{308}   & \num{7}  & Regression \\
\bottomrule
\end{tabular}
\end{table}

\begin{table}[h]
\centering
\caption{Baseline default parameter settings of \textsc{TabINR}.}
\label{tab:Tab5}
\begin{tabular}{l|c}
\toprule
\textbf{Parameter} & \textbf{Setting} \\
\midrule
Latent dimension              & \num{32} \\
Number of hidden layers       & \num{2} \\
Number of units per hidden layer & \num{256} \\
Dropout rate                  & \num{0.1} \\
Activation function           & SIREN \\
$\omega_0$ (for SIREN, Wire activations) & \num{30} \\
Learning rate                 & \num{1e-3} \\
Training epochs               & \num{500} \\
Optimizer                     & Adam \\
Batch size                    & \num{64} \\
Masking ratio                 & \num{0.3} \\
\bottomrule
\end{tabular}
\end{table}

\begin{figure}[t]
    \centering
    \includegraphics[width=1.0\textwidth]{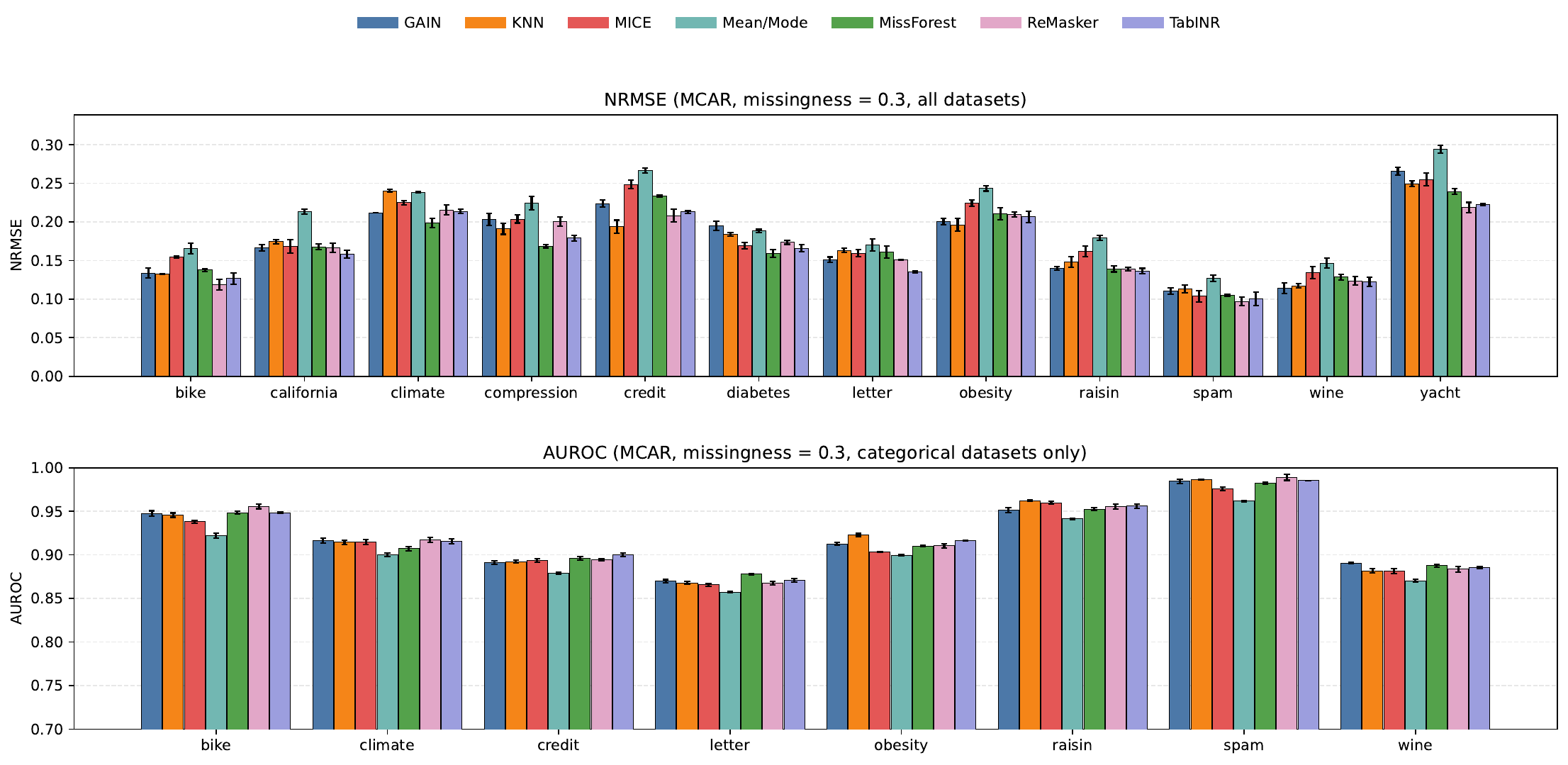}
    \caption{Overall performance of \textsc{TabINR} and six baselines on \num{12} benchmark datasets under \texttt{MCAR} with
\num{0.3} missingness ratio. The results are shown as the mean and standard deviation of NRMSE, and
AUROC scores (AUROC is only applicable to datasets with classification tasks).}
\label{fig:Fig5}
\end{figure}

\begin{figure}[t]
    \centering
    \includegraphics[width=1.0\textwidth]{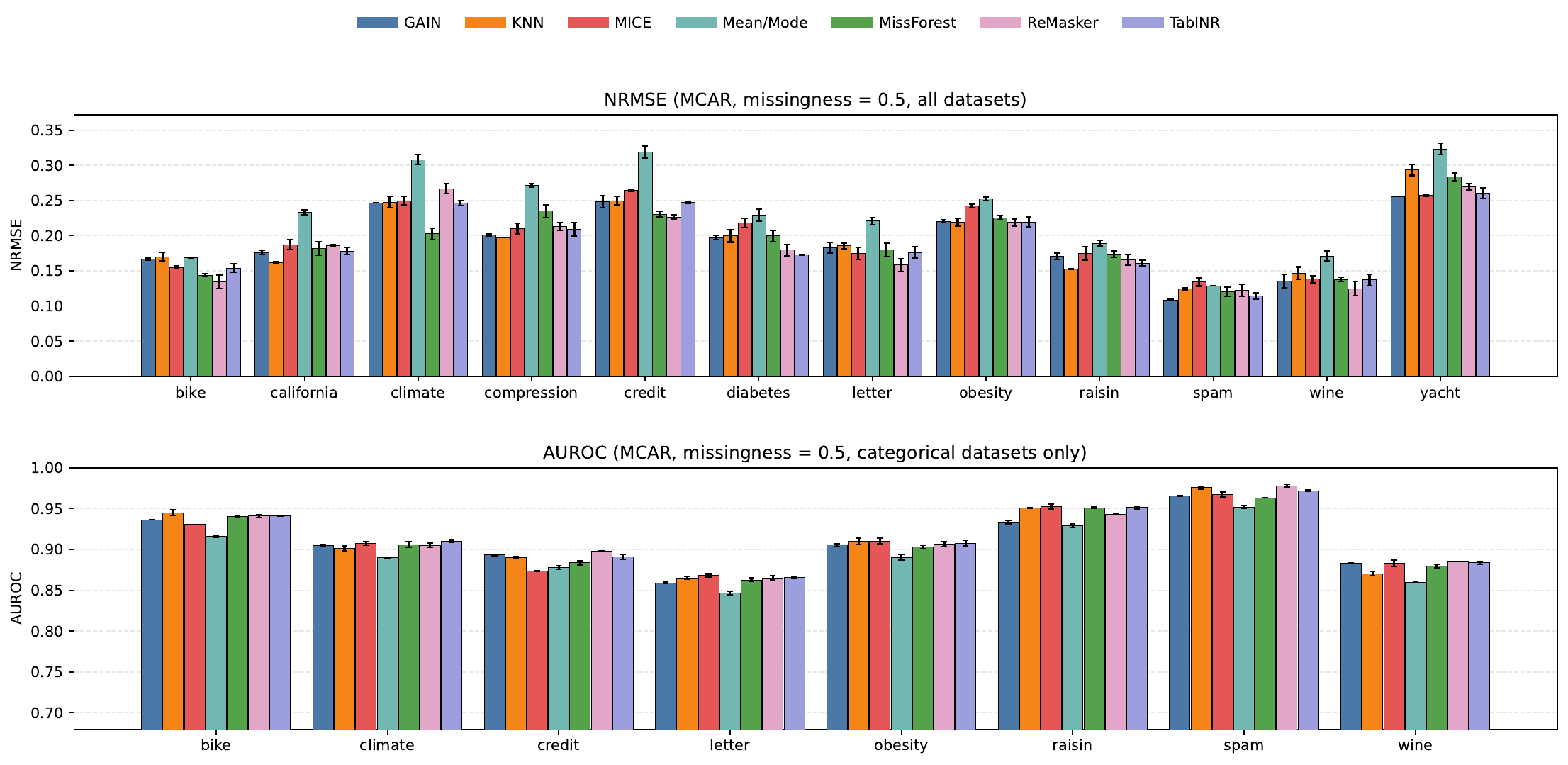}
    \caption{Overall performance of \textsc{TabINR} and six baselines on \num{12} benchmark datasets under \texttt{MCAR} with
\num{0.5} missingness ratio. The results are shown as the mean and standard deviation of NRMSE, and
AUROC scores (AUROC is only applicable to datasets with classification tasks).}
\label{fig:Fig6}
\end{figure}

\begin{figure}[t]
    \centering
    \includegraphics[width=1.0\textwidth]{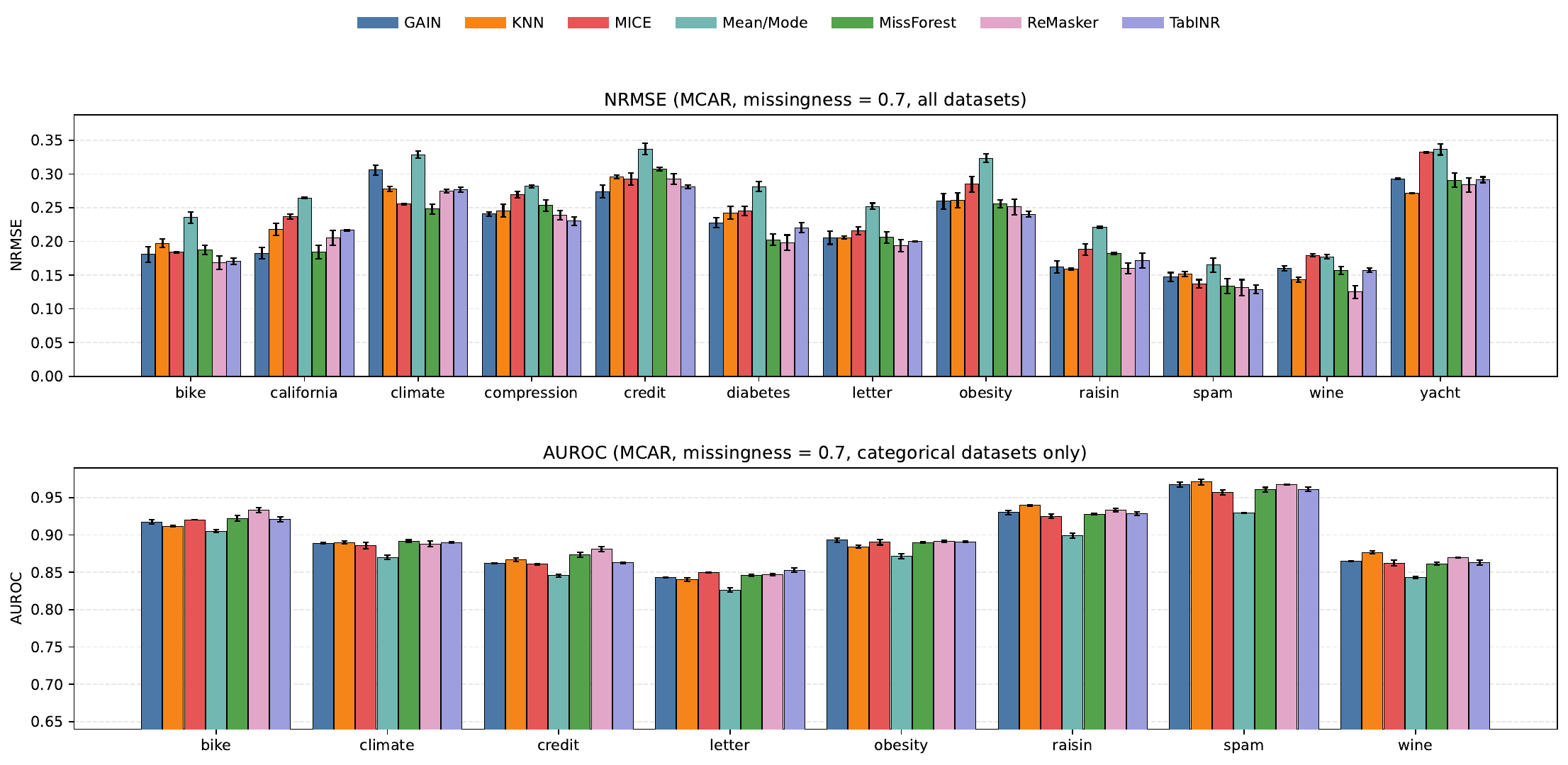}
    \caption{Overall performance of \textsc{TabINR} and six baselines on \num{12} benchmark datasets under \texttt{MCAR} with
\num{0.7} missingness ratio. The results are shown as the mean and standard deviation of NRMSE, and
AUROC scores (AUROC is only applicable to datasets with classification tasks).}
\label{fig:Fig7}
\end{figure}

\begin{figure}[t]
    \centering
    \includegraphics[width=1.0\textwidth]{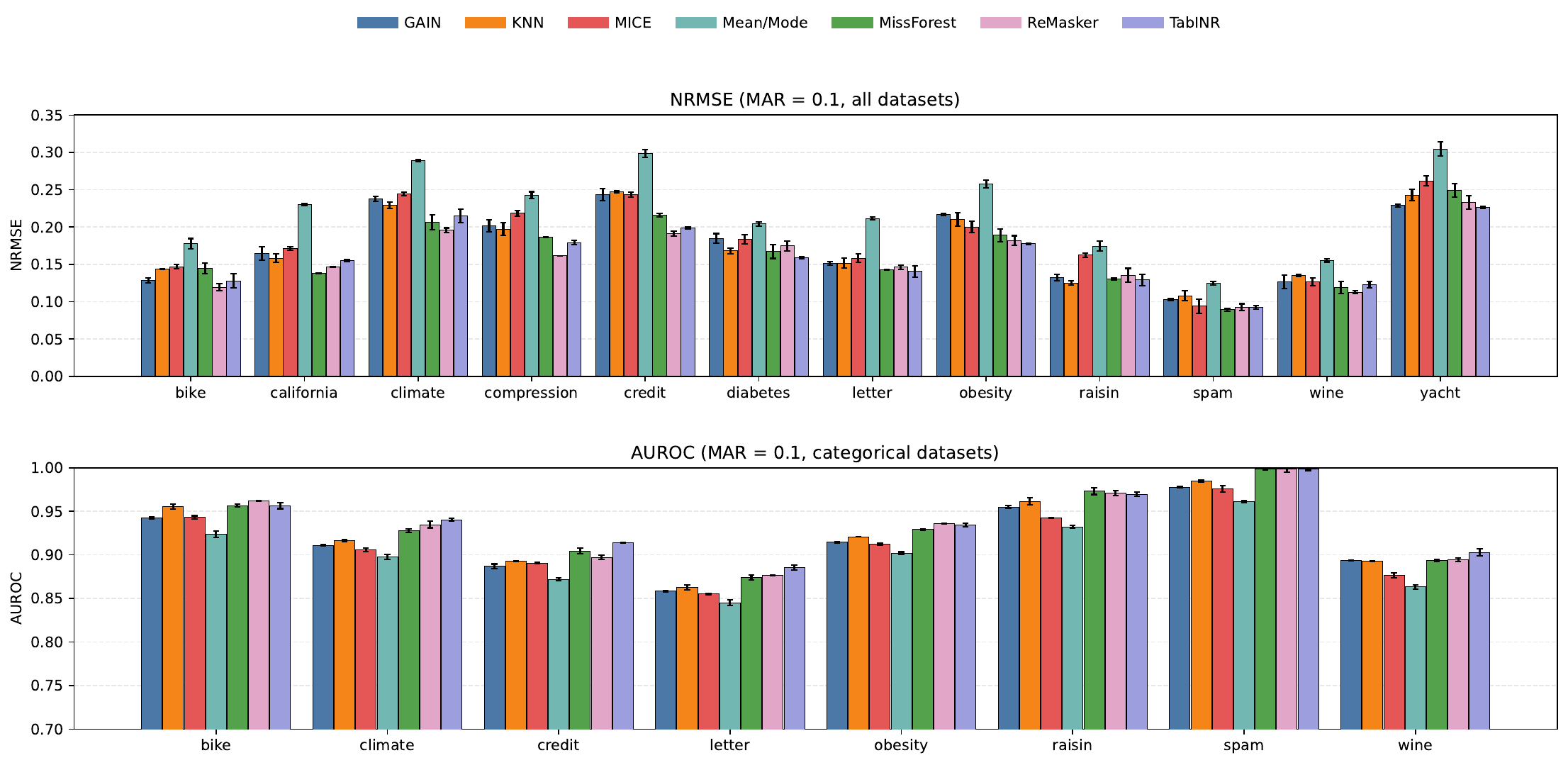}
    \caption{Overall performance of \textsc{TabINR} and six baselines on \num{12} benchmark datasets under \texttt{MAR} with
\num{0.1} missingness ratio. The results are shown as the mean and standard deviation of NRMSE, and
AUROC scores (AUROC is only applicable to datasets with classification tasks).}
\label{fig:Fig8}
\end{figure}

\begin{figure}[t]
    \centering
    \includegraphics[width=1.0\textwidth]{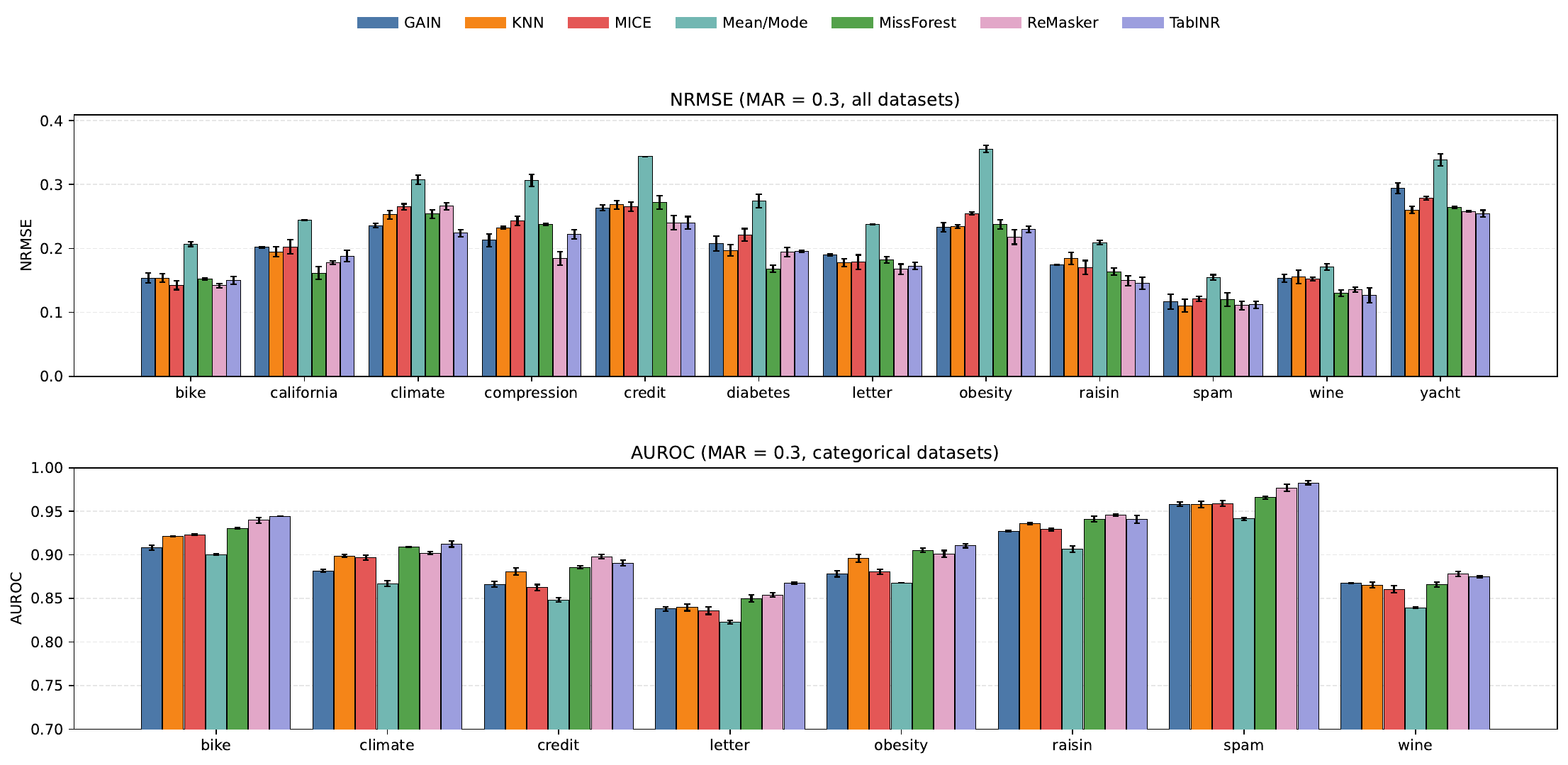}
    \caption{Overall performance of \textsc{TabINR} and six baselines on \num{12} benchmark datasets under \texttt{MAR} with
\num{0.3} missingness ratio. The results are shown as the mean and standard deviation of NRMSE, and
AUROC scores (AUROC is only applicable to datasets with classification tasks).}
\label{fig:Fig9}
\end{figure}

\begin{figure}[t]
    \centering
    \includegraphics[width=1.0\textwidth]{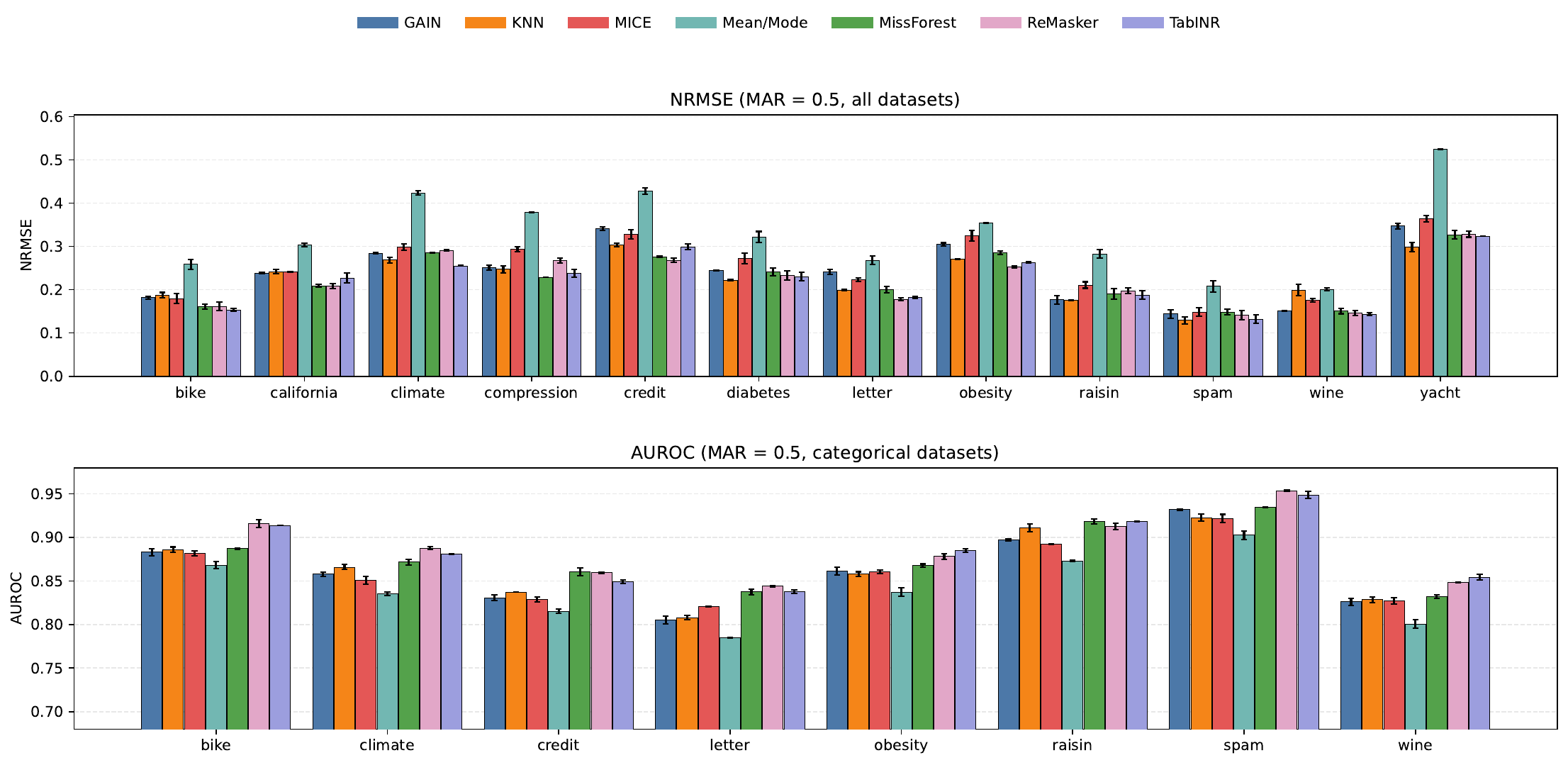}
    \caption{Overall performance of \textsc{TabINR} and six baselines on \num{12} benchmark datasets under \texttt{MAR} with
\num{0.5} missingness ratio. The results are shown as the mean and standard deviation of RMSE, and
AUROC scores (AUROC is only applicable to datasets with classification tasks).}
\label{fig:Fig10}
\end{figure}

\begin{figure}[t]
    \centering
    \includegraphics[width=1.0\textwidth]{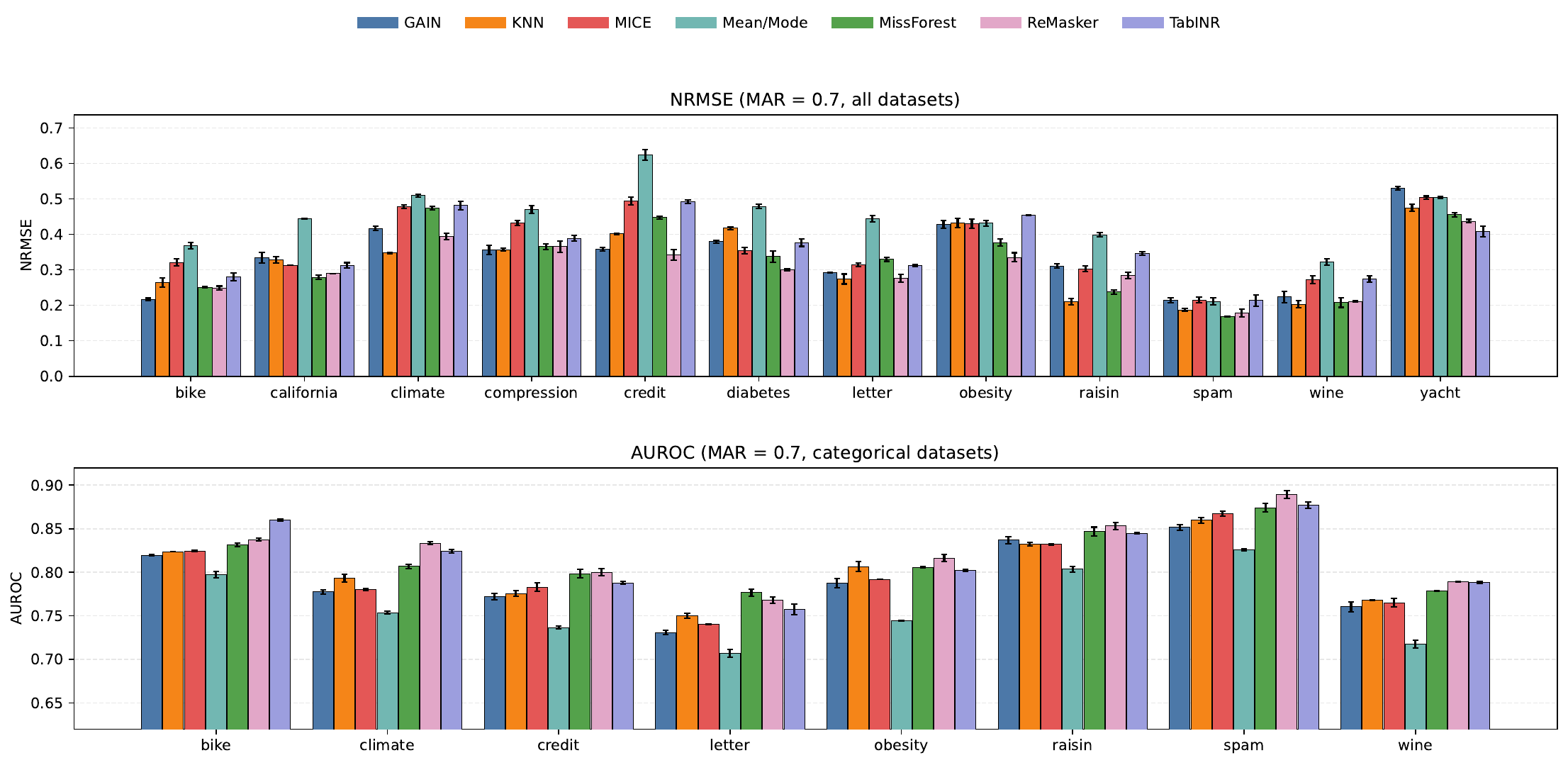}
    \caption{Overall performance of \textsc{TabINR} and six baselines on \num{12} benchmark datasets under \texttt{MAR} with
\num{0.7} missingness ratio. The results are shown as the mean and standard deviation of NRMSE, and
AUROC scores (AUROC is only applicable to datasets with classification tasks).}
\label{fig:Fig11}
\end{figure}

\begin{figure}[t]
    \centering
    \includegraphics[width=1.0\textwidth]{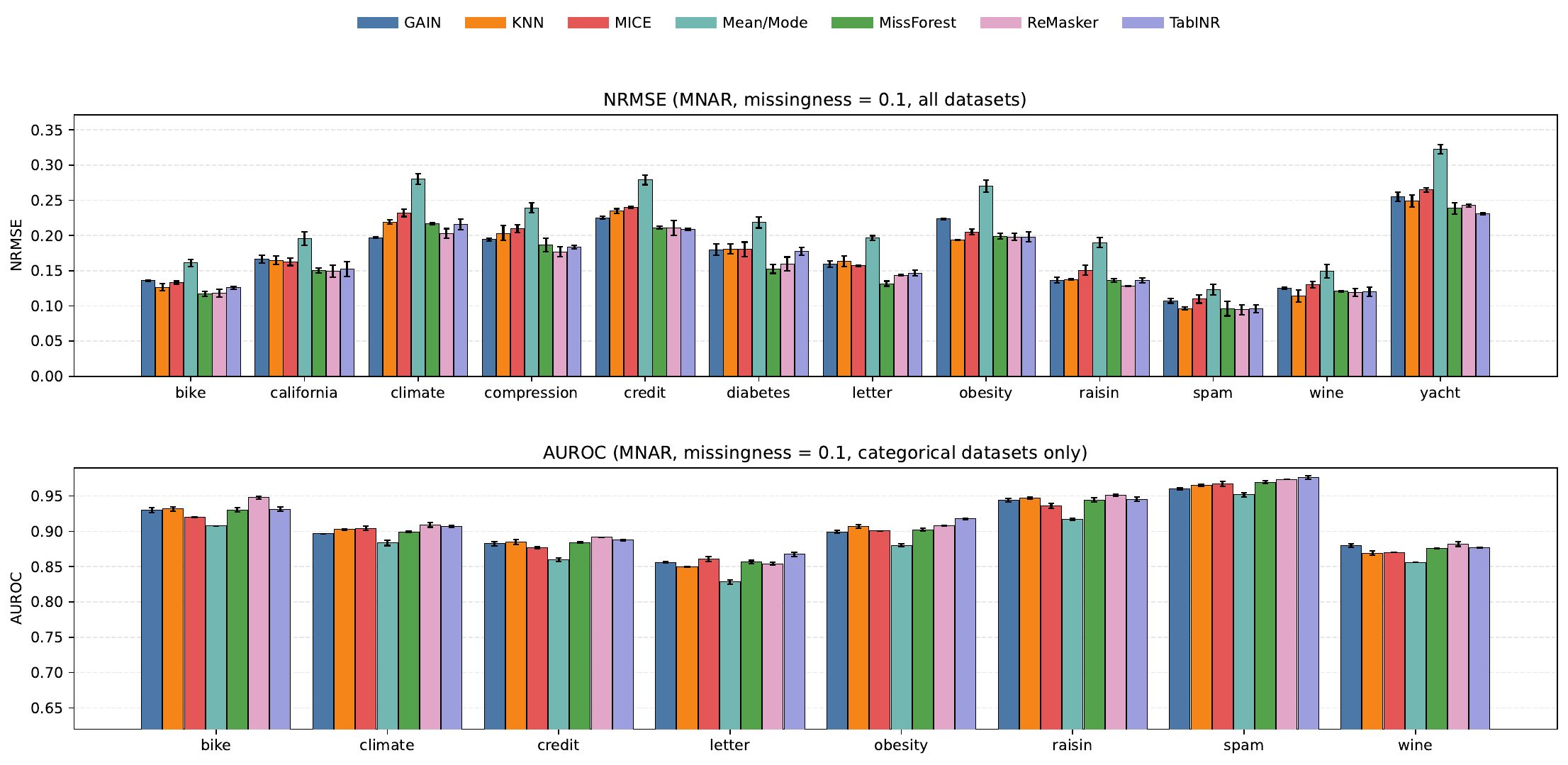}
    \caption{Overall performance of \textsc{TabINR} and six baselines on \num{12} benchmark datasets under \texttt{MNAR} with
\num{0.1} missingness ratio. The results are shown as the mean and standard deviation of NRMSE, and
AUROC scores (AUROC is only applicable to datasets with classification tasks).}
\label{fig:Fig12}
\end{figure}

\begin{figure}[t]
    \centering
    \includegraphics[width=1.0\textwidth]{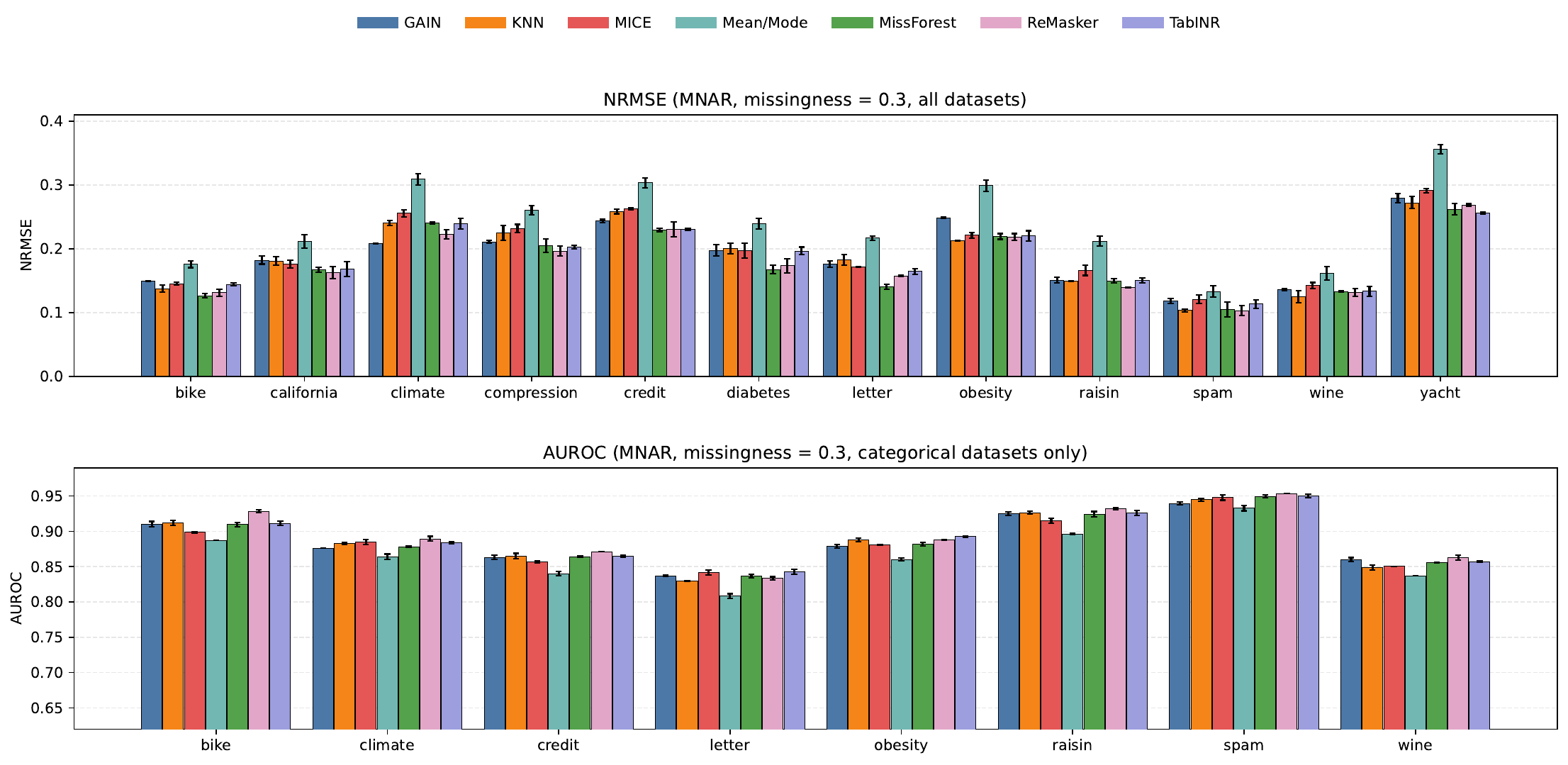}
     \caption{Overall performance of \textsc{TabINR} and six baselines on \num{12} benchmark datasets under \texttt{MNAR} with
\num{0.3} missingness ratio. The results are shown as the mean and standard deviation of NRMSE, and
AUROC scores (AUROC is only applicable to datasets with classification tasks).}
\label{fig:Fig13}
\end{figure}

\begin{figure}[t]
    \centering
    \includegraphics[width=1.0\textwidth]{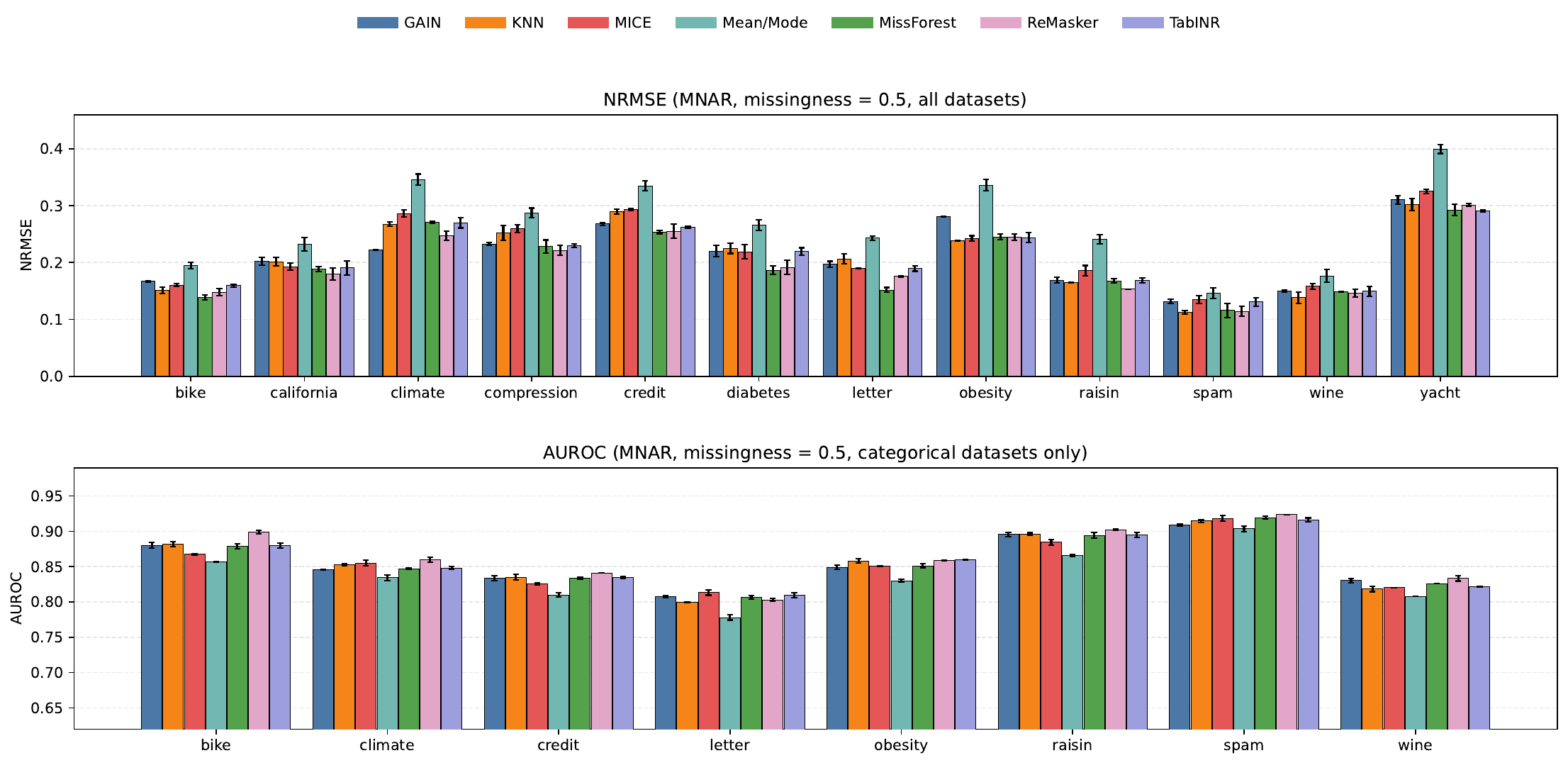}
    \caption{Overall performance of \textsc{TabINR} and six baselines on \num{12} benchmark datasets under \texttt{MNAR} with
\num{0.5} missingness ratio. The results are shown as the mean and standard deviation of NRMSE, and
AUROC scores (AUROC is only applicable to datasets with classification tasks).}
\label{fig:Fig14}
\end{figure}

\begin{figure}[t]
    \centering
    \includegraphics[width=1.0\textwidth]{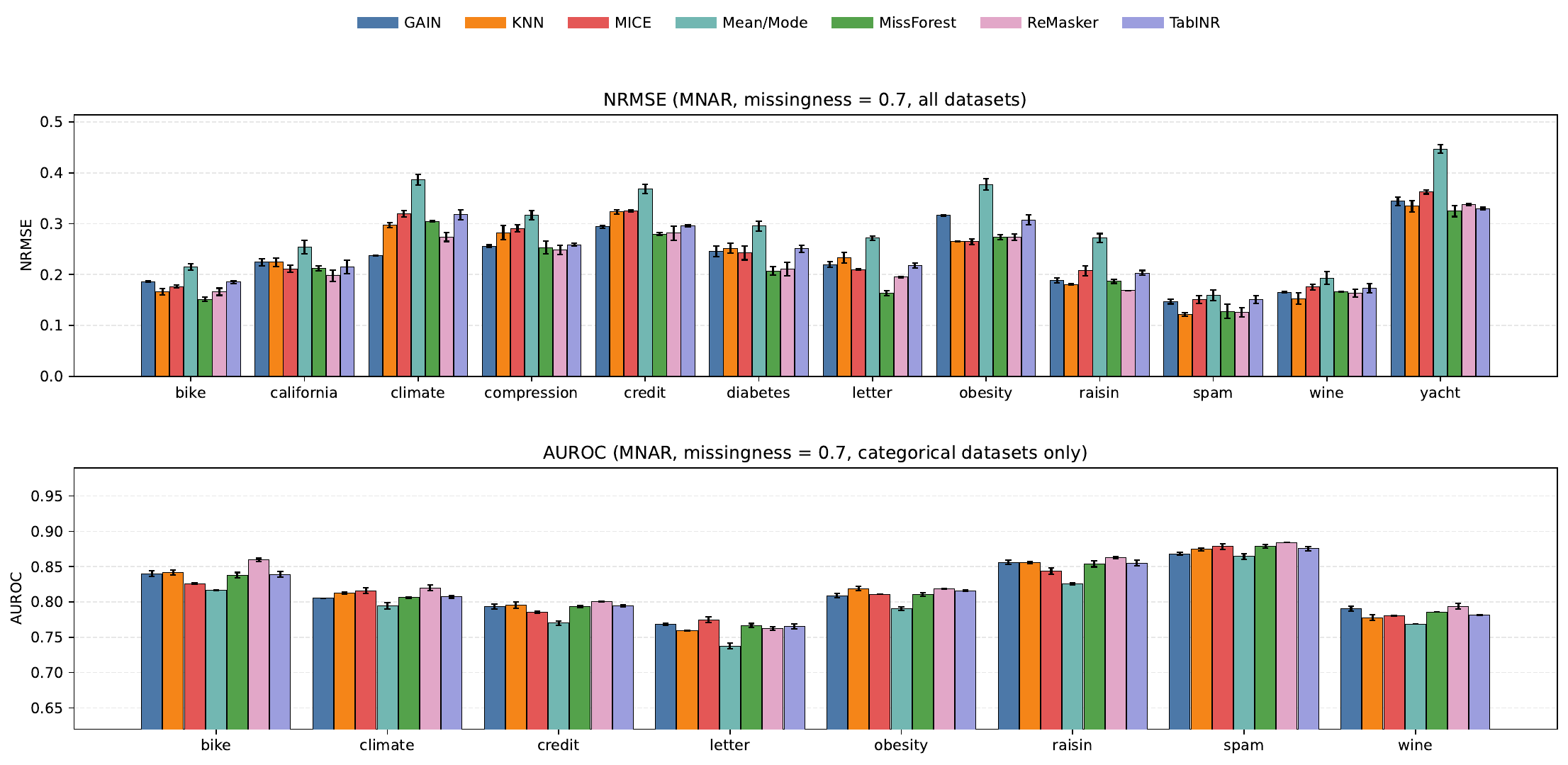}
    \caption{Overall performance of \textsc{TabINR} and six baselines on \num{12} benchmark datasets under \texttt{MNAR} with
\num{0.7} missingness ratio. The results are shown as the mean and standard deviation of NRMSE, and
AUROC scores (AUROC is only applicable to datasets with classification tasks).}
\label{fig:Fig15}
\end{figure}

\begin{figure}[t]
    \centering
    \includegraphics[width=1.0\textwidth]{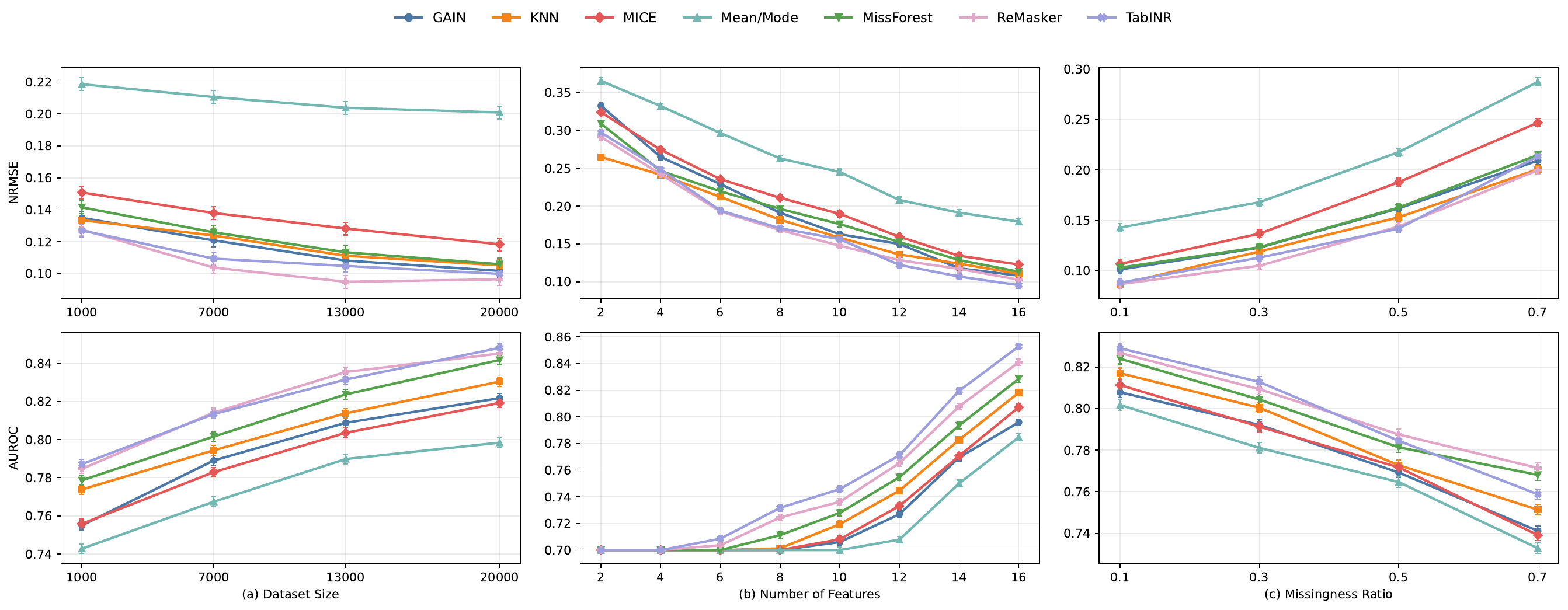}
    \caption{Sensitivity analysis of \textsc{TabINR} on the \textit{letter} dataset under \texttt{MCAR} scenarios.
The results are shown in terms of NRMSE and AUROC, with the scores measured with respect to (a) the dataset size,
(b) the number of features, and (c) the missingness ratio. The default setting is as follows: dataset size = \num{20000},
number of features = \num{16}, and missingness ratio = \num{0.3}.}
\label{fig:Fig16}
\end{figure}

\begin{figure}[t]
    \centering
    \includegraphics[width=1.0\textwidth]{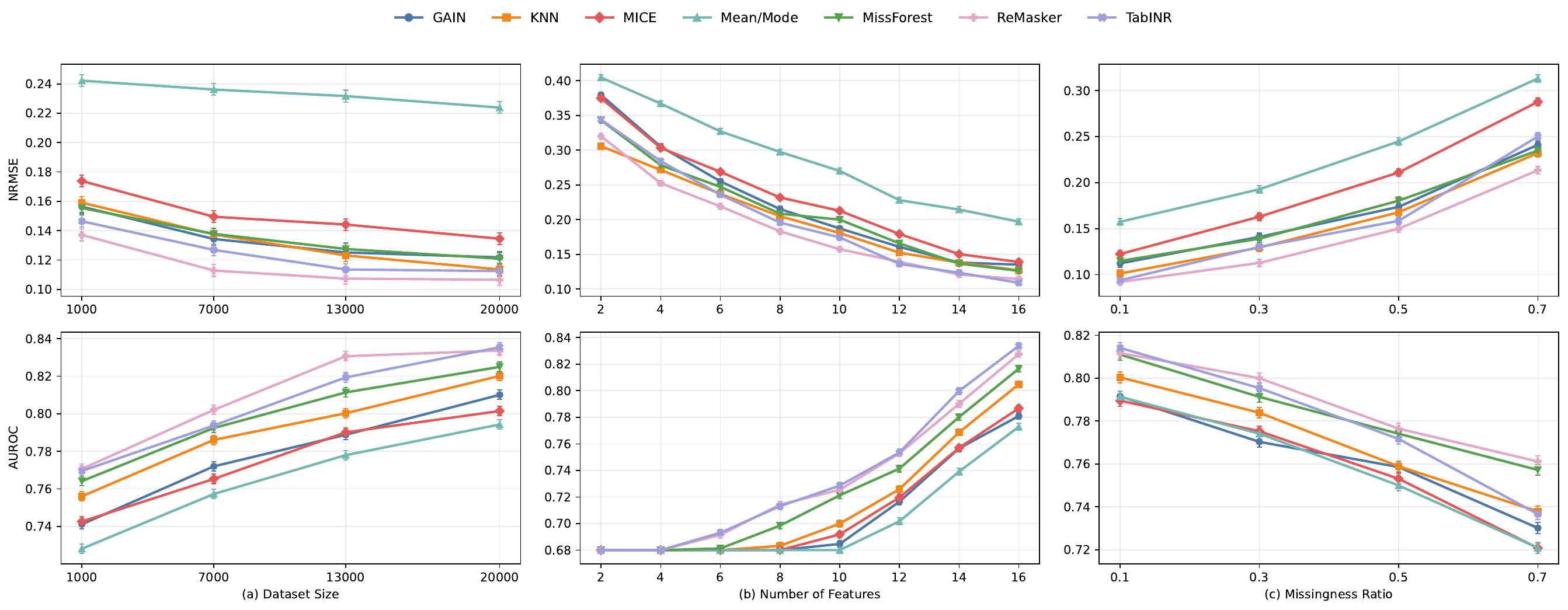}
    \caption{Sensitivity analysis of \textsc{TabINR} on the \textit{letter} dataset under \texttt{MCAR} scenarios.
The results are shown in terms of NRMSE and AUROC, with the scores measured with respect to (a) the dataset size,
(b) the number of features, and (c) the missingness ratio. The default setting is as follows: dataset size = \num{20000},
number of features = \num{16}, and missingness ratio = \num{0.5}.}
\label{fig:Fig17}
\end{figure}

\begin{figure}[t]
    \centering
    \includegraphics[width=1.0\textwidth]{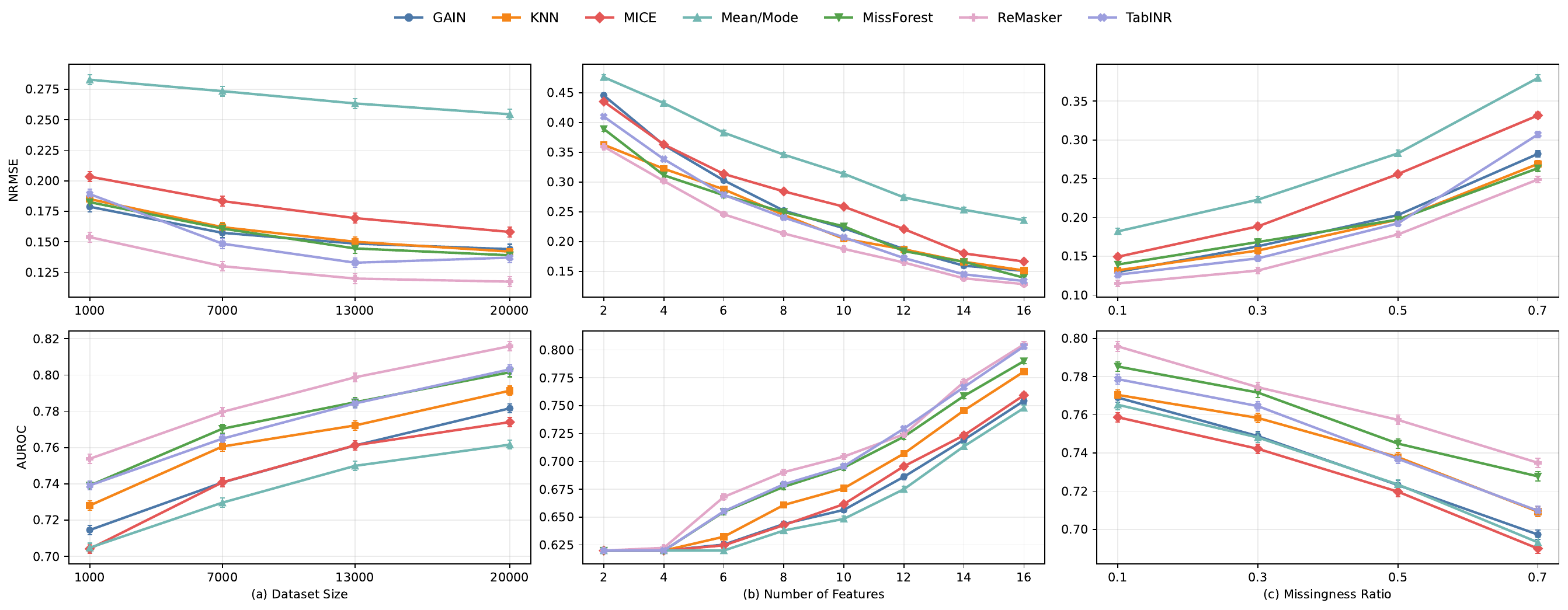}
    \caption{Sensitivity analysis of \textsc{TabINR} on the \textit{letter} dataset under \texttt{MCAR} scenarios.
The results are shown in terms of NRMSE and AUROC, with the scores measured with respect to (a) the dataset size,
(b) the number of features, and (c) the missingness ratio. The default setting is as follows: dataset size = \num{20000},
number of features = \num{16}, and missingness ratio = \num{0.7}.}
\label{fig:Fig18}
\end{figure}

\begin{figure}[t]
    \centering
    \includegraphics[width=1.0\textwidth]{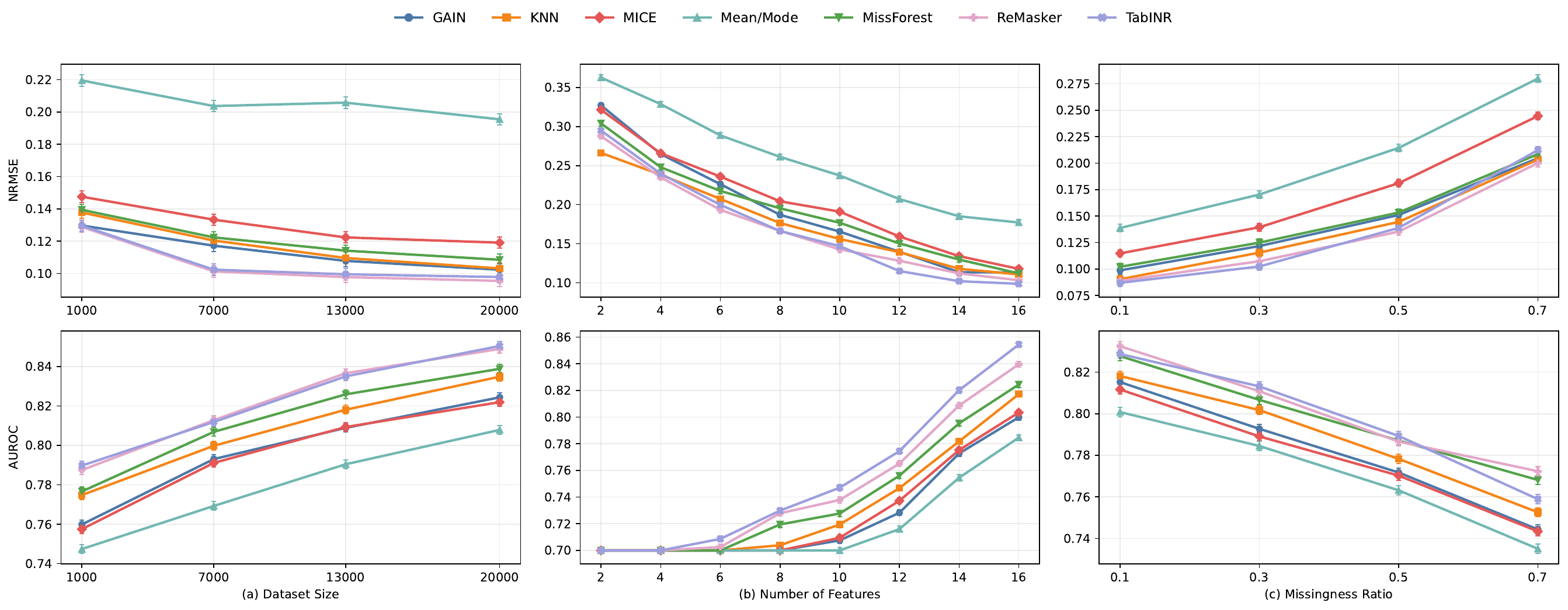}
    \caption{Sensitivity analysis of \textsc{TabINR} on the \textit{letter} dataset under \texttt{MAR} scenarios.
The results are shown in terms of NRMSE and AUROC, with the scores measured with respect to (a) the dataset size,
(b) the number of features, and (c) the missingness ratio. The default setting is as follows: dataset size = \num{20000},
number of features = \num{16}, and missingness ratio = \num{0.1}.}
\label{fig:Fig19}
\end{figure}

\begin{figure}[t]
    \centering
    \includegraphics[width=1.0\textwidth]{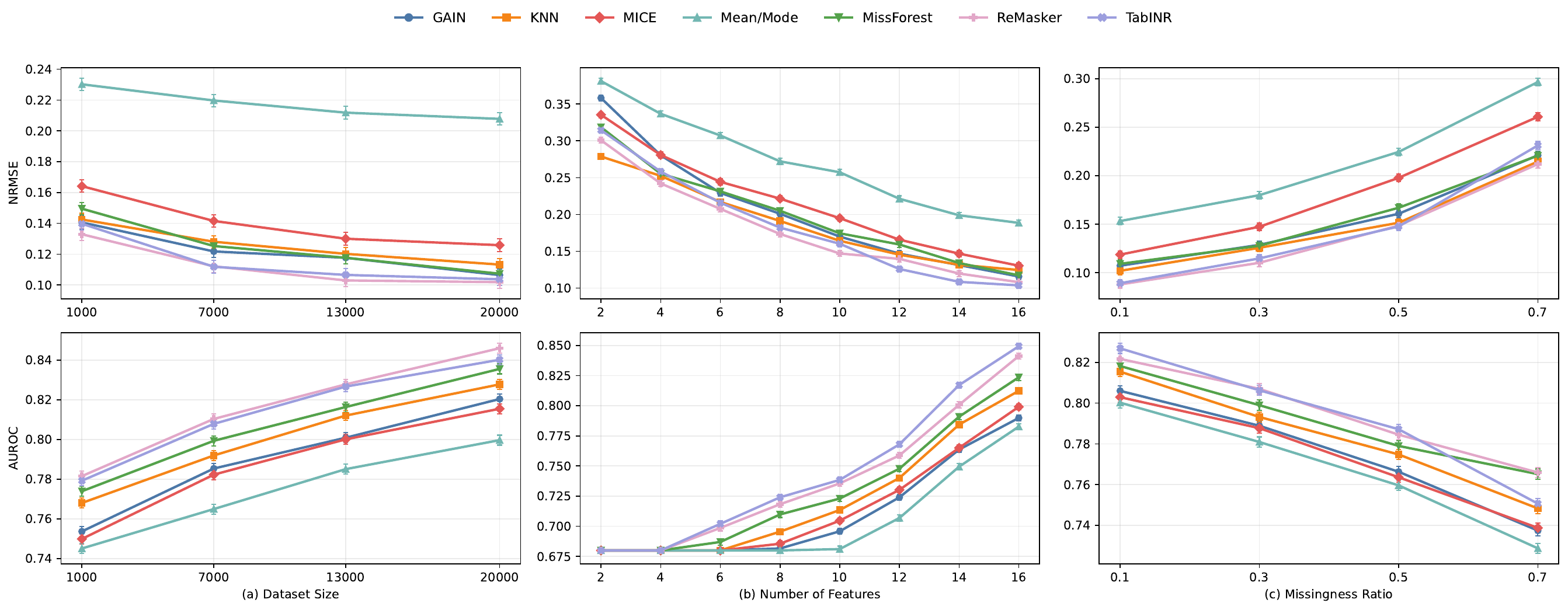}
    \caption{Sensitivity analysis of \textsc{TabINR} on the \textit{letter} dataset under \texttt{MAR} scenarios.
The results are shown in terms of NRMSE and AUROC, with the scores measured with respect to (a) the dataset size,
(b) the number of features, and (c) the missingness ratio. The default setting is as follows: dataset size = \num{20000},
number of features = \num{16}, and missingness ratio = \num{0.3}.}
\label{fig:Fig20}
\end{figure}

\begin{figure}[t]
    \centering
    \includegraphics[width=1.0\textwidth]{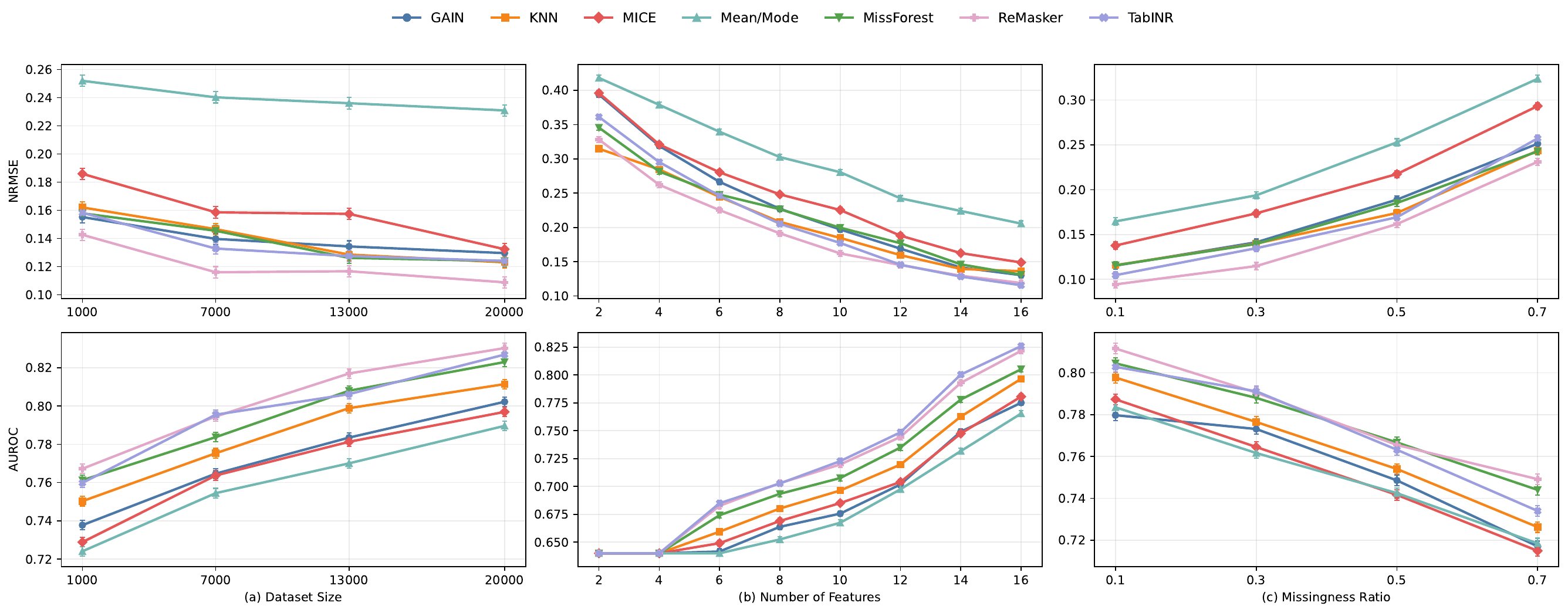}
    \caption{Sensitivity analysis of \textsc{TabINR} on the \textit{letter} dataset under \texttt{MAR} scenarios.
The results are shown in terms of NRMSE and AUROC, with the scores measured with respect to (a) the dataset size,
(b) the number of features, and (c) the missingness ratio. The default setting is as follows: dataset size = \num{20000},
number of features = \num{16}, and missingness ratio = \num{0.5}.}
\label{fig:Fig21}
\end{figure}

\begin{figure}[t]
    \centering
    \includegraphics[width=1.0\textwidth]{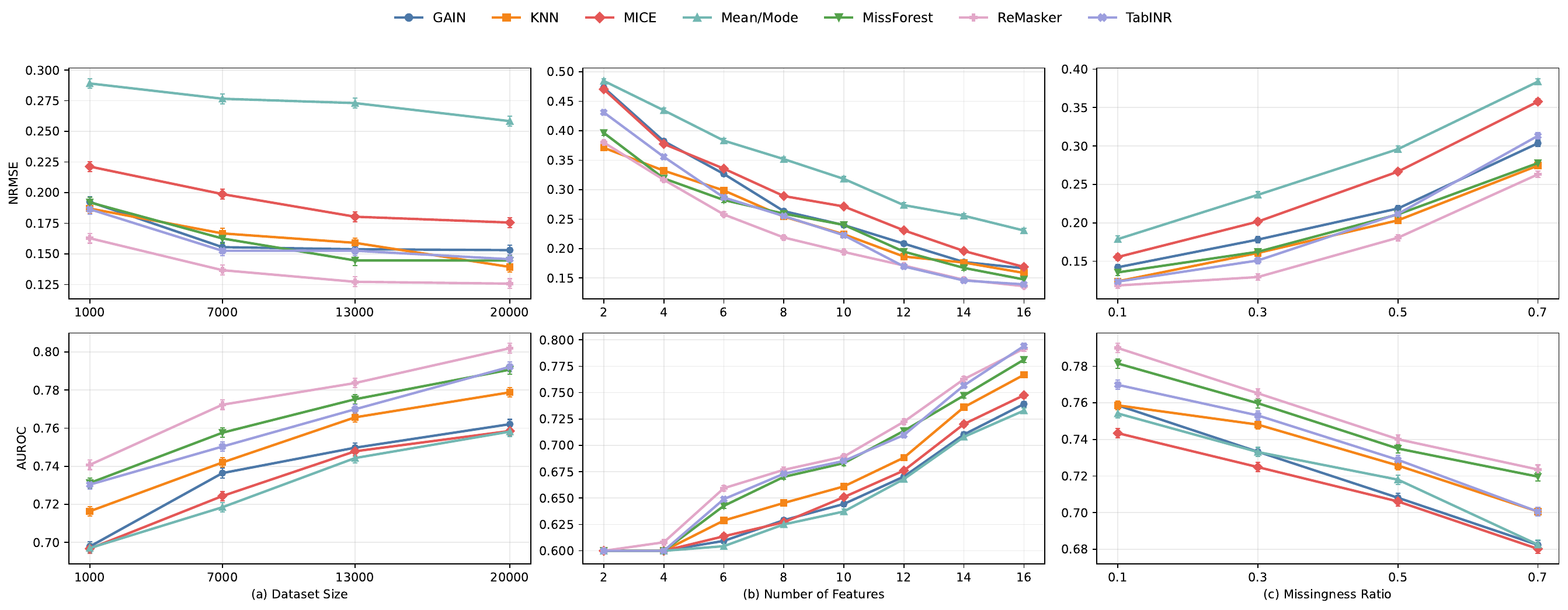}
    \caption{Sensitivity analysis of \textsc{TabINR} on the \textit{letter} dataset under \texttt{MAR} scenarios.
The results are shown in terms of NRMSE and AUROC, with the scores measured with respect to (a) the dataset size,
(b) the number of features, and (c) the missingness ratio. The default setting is as follows: dataset size = \num{20000},
number of features = \num{16}, and missingness ratio = \num{0.7}.}
\label{fig:Fig22}
\end{figure}

\begin{figure}[t]
    \centering
    \includegraphics[width=1.0\textwidth]{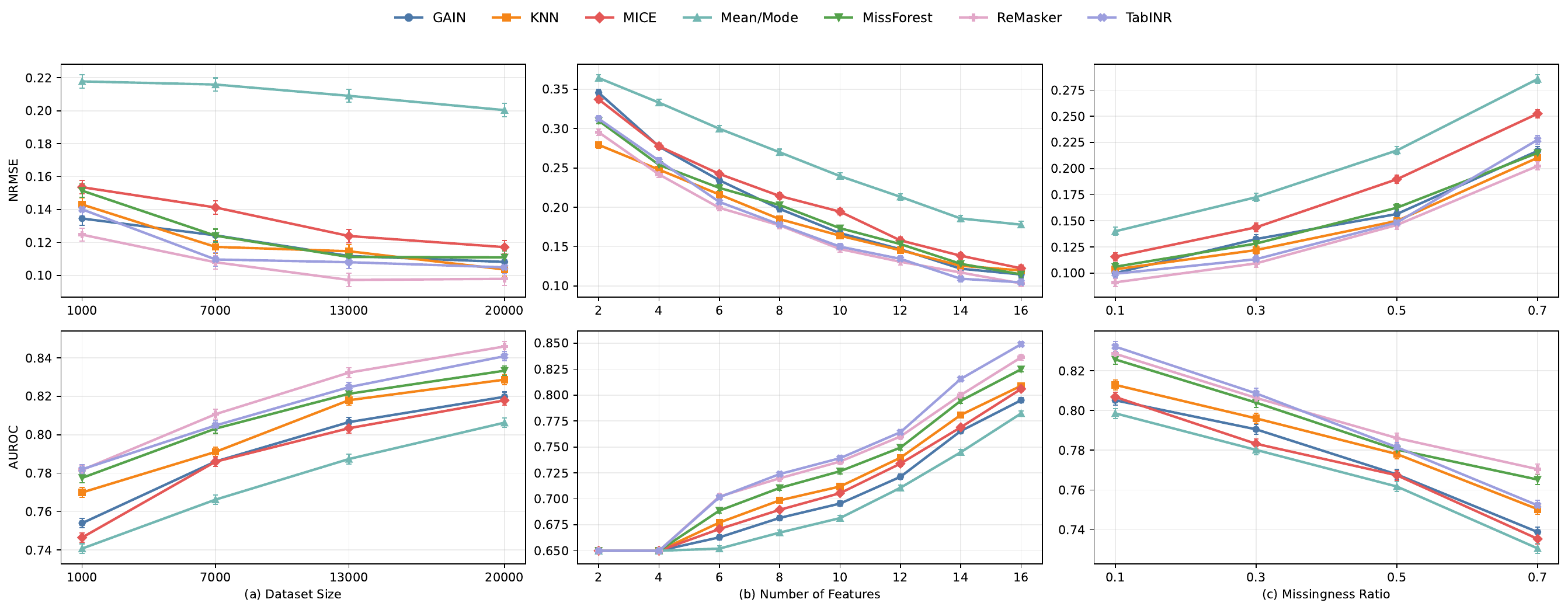}
    \caption{Sensitivity analysis of \textsc{TabINR} on the \textit{letter} dataset under \texttt{MNAR} scenarios.
The results are shown in terms of NRMSE and AUROC, with the scores measured with respect to (a) the dataset size,
(b) the number of features, and (c) the missingness ratio. The default setting is as follows: dataset size = \num{20000},
number of features = \num{16}, and missingness ratio = \num{0.1}.}
\label{fig:Fig23}
\end{figure}

\begin{figure}[t]
    \centering
    \includegraphics[width=1.0\textwidth]{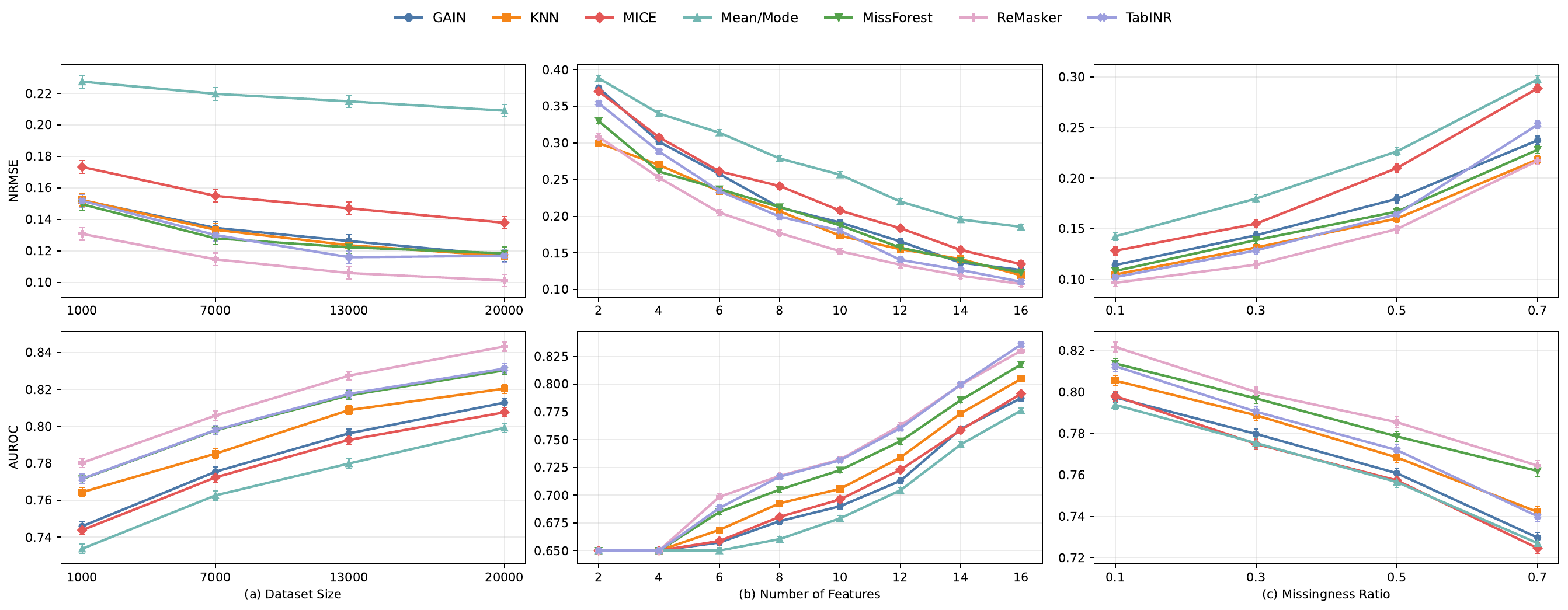}
    \caption{Sensitivity analysis of \textsc{TabINR} on the \textit{letter} dataset under \texttt{MNAR} scenarios.
The results are shown in terms of NRMSE and AUROC, with the scores measured with respect to (a) the dataset size,
(b) the number of features, and (c) the missingness ratio. The default setting is as follows: dataset size = \num{20000},
number of features = \num{16}, and missingness ratio = \num{0.3}.}
\label{fig:Fig24}
\end{figure}

\begin{figure}[t]
    \centering
    \includegraphics[width=1.0\textwidth]{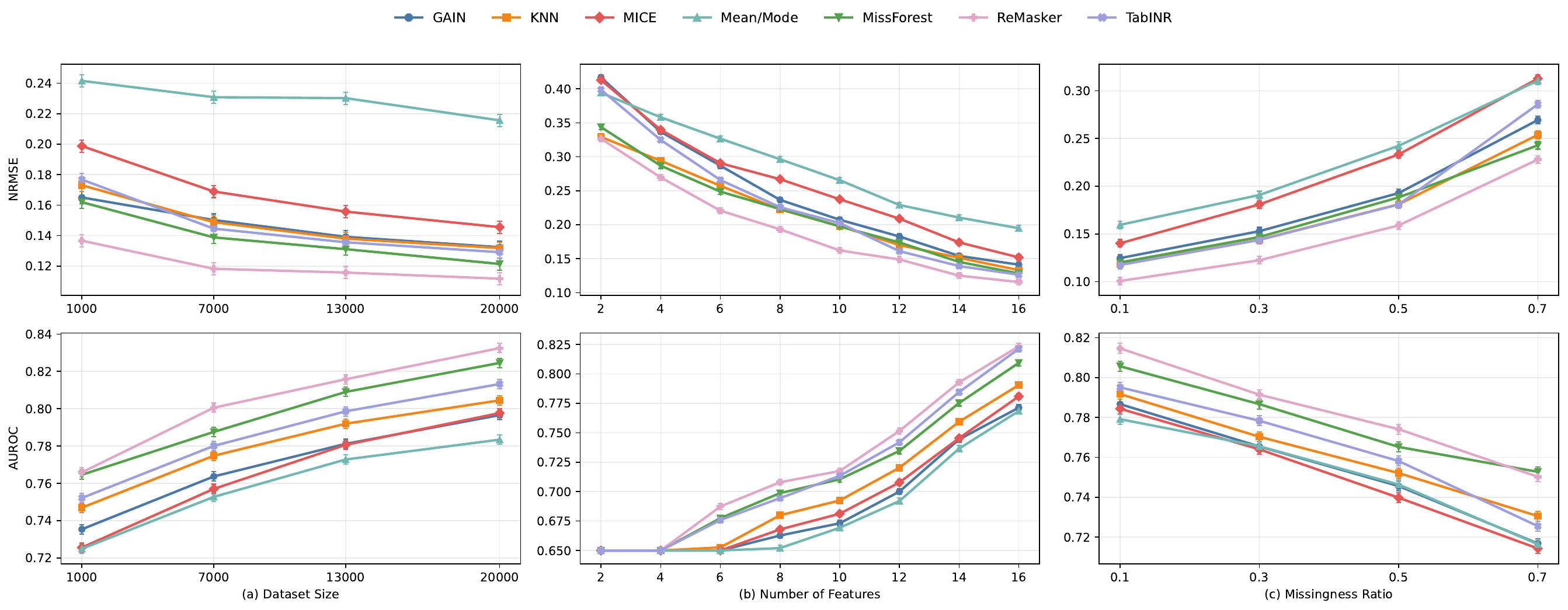}
    \caption{Sensitivity analysis of \textsc{TabINR} on the \textit{letter} dataset under \texttt{MNAR} scenarios.
The results are shown in terms of NRMSE and AUROC, with the scores measured with respect to (a) the dataset size,
(b) the number of features, and (c) the missingness ratio. The default setting is as follows: dataset size = \num{20000},
number of features = \num{16}, and missingness ratio = \num{0.5}.}
\label{fig:Fig25}
\end{figure}

\begin{figure}[t]
    \centering
    \includegraphics[width=1.0\textwidth]{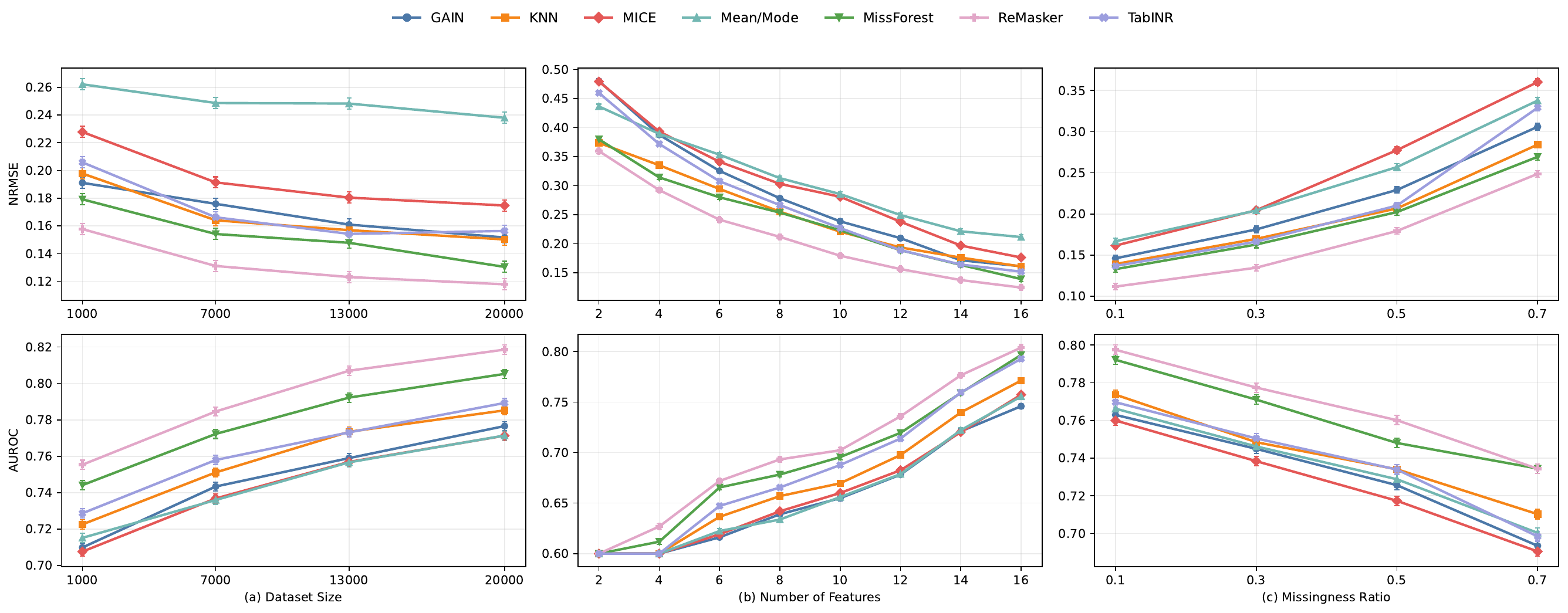}
    \caption{: Sensitivity analysis of \textsc{TabINR} on the \textit{letter} dataset under \texttt{MNAR} scenarios.
The results are shown in terms of NRMSE and AUROC, with the scores measured with respect to (a) the dataset size,
(b) the number of features, and (c) the missingness ratio. The default setting is as follows: dataset size = \num{20000},
number of features = \num{16}, and missingness ratio = \num{0.7}.}
\label{fig:Fig26}
\end{figure}

\end{document}